\documentclass[dvipsnames,format=sigconf]{acmart}

\AtBeginDocument{%
  }

\copyrightyear{2026}
\acmYear{2026}
\setcopyright{cc}
\setcctype{by}
\acmConference[GECCO Companion '26]{Genetic and Evolutionary Computation Conference}{July 13--17, 2026}{San Jose, Costa Rica}
\acmBooktitle{Genetic and Evolutionary Computation Conference (GECCO Companion '26), July 13--17, 2026, San Jose, Costa Rica}
\acmDOI{10.1145/3795101.3814681}
\acmISBN{979-8-4007-2488-6/2026/07}

\usepackage{multirow}
\usepackage{enumitem}
\usepackage{pdfpages}

\begin{document}


\title{Beyond the Training Distribution: Mapping Generalization Boundaries in Neural Program Synthesis}





\author{Henrik Voigt}
\email{henrik.voigt@uni-jena.de}
\orcid{0009-0003-3671-4582}
\affiliation{%
  \institution{Friedrich Schiller University}
  \city{Jena}
  \country{Germany}
}

\author{Michael Habeck}
\email{michael.habeck@uni-jena.de}
\orcid{0000-0002-2188-5667}
\affiliation{%
  \institution{Friedrich Schiller University}
  \city{Jena}
  \country{Germany}
}

\author{Joachim Giesen}
\email{joachim.giesen@uni-jena.de}
\orcid{0000-0001-6598-6833}
\affiliation{%
  \institution{Friedrich Schiller University}
  \city{Jena}
  \country{Germany}
}

\renewcommand{\shortauthors}{Trovato et al.}


\begin{abstract}
Large-scale transformers achieve impressive results on program synthesis benchmarks, yet their true generalization capabilities remain obscured by data contamination and opaque training corpora. To rigorously assess whether models are truly generalizing or merely retrieving memorized templates, we introduce a strictly controlled program synthesis environment based on a domain-specific arithmetic grammar. By systematically enumerating and evaluating millions of unique programs, we construct interpretable syntactic and semantic metric spaces. This allows us to precisely map data distributions and sample train and test splits that isolate specific distributional shifts. Our experiments demonstrate that optimizing \textit{density generalization}---through diverse sampling over both semantic and syntactic spaces---induces robust out-of-distribution generalization. Conversely, evaluating \textit{support generalization} reveals that transformers severely struggle with extrapolation, experiencing a performance drop of over 30\% when forced to generate syntactically novel programs. While steadily scaling up compute improves generalization, the gains follow a strictly log-linear relationship. We conclude that robust generalization requires maximizing training diversity across multiple manifolds, and our findings indicate the necessity for novel search-based approaches to break through current log-linear scaling bottlenecks.
\end{abstract}


  



\maketitle

\section{Introduction}
\label{sec:introduction}

The ability to efficiently generalize from a few examples and apply this generalized knowledge to novel problems is a hallmark of human intelligence~\cite{lake2015human}. This trait is particularly crucial when inventing a novel algorithmic solution, such as in a program synthesis task. At first glance, recent large-scale transformer language models exhibit impressive code generation capabilities across various benchmarks~\cite{jiang2026survey}. 

However, because the training corpora of these frontier models---even open-weight ones---are often kept proprietary or remain highly opaque, it is extremely difficult to develop a rigorous understanding of their true generalization capabilities. Without knowing what a model has seen during training, it is nearly impossible to ascertain whether it is genuinely synthesizing novel solutions or merely retrieving and interpolating memorized templates. 

Recent studies, such as the work by~\citet{riddell2024quantifying}, have revealed severe data contamination in prominent benchmarks like MBPP~\cite{austin2021program} and HumanEval~\cite{chen2021evaluating}. These overlaps between training and test distributions artificially inflate performance metrics. To mitigate this, live benchmarks \cite{jain2024livecodebench} use post-training cutoff problems (e.g., recent competitive programming tasks from LeetCode). While promising, these benchmarks still fail to provide researchers with the full investigative power needed to understand generalization boundaries. Without access to the pre-training data, it is impossible to measure how far ``out-of-distribution'' (OOD) a novel test instance truly is. Consequently, we cannot systematically evaluate the limits of a transformer's generalization capabilities.

For the broader program synthesis and evolutionary computation communities, mapping these boundaries is critical. If transformers are largely limited to local interpolation within their training distribution, building robust, general-purpose program synthesis systems will increasingly require combining neural models with search-based or genetic algorithms to successfully extrapolate into unexplored regions of the program space.

To systematically quantify these boundaries, we design a strictly controlled program synthesis environment. We define a context-free grammar of arithmetic programs and methodically enumerate millions of problem instances. By computing their syntactic and semantic similarities, we project these programs into continuous syntactic and semantic metric spaces. This controlled setup allows us to precisely map the data distribution and systematically derive in-distribution and out-of-distribution train and test splits, where the geometric distance between any test instance and the training set is known.

Using this framework, we investigate two distinct modes of generalization: \textit{density generalization} (shifts in the data distribution within the same support) and \textit{support generalization} (extrapolation to regions entirely outside the training support). We find that moving OOD in syntactic space degrades the \textit{pass@k} performance of transformer models significantly. Surprisingly, we demonstrate that a deliberate selection of pre-training instances---sampled diversely over the semantic and syntactic space rather than drawn from a single distribution, e.g., maximizing syntactic diversity only---yields vastly superior generalization, offering crucial insights for the practical implementation of program synthesis models. 

In summary, our core contributions are:
\begin{itemize}
    \item \textbf{A controlled evaluation framework:} We introduce a fully transparent, grammar-based program synthesis task paired with continuous syntactic and semantic manifolds to formally quantify OOD distances.
    \item \textbf{Generalization taxonomy:} We formalize and evaluate two distinct generalization modalities in neural program synthesis: \textit{density generalization} and \textit{support generalization}.
    \item \textbf{Data distribution insights:} We demonstrate empirically that diversely sampling training data across both the semantic and syntactic space effectively increases OOD generalization performance (from ${\sim}10\%$ to ${\sim}19\%$ pass@1) compared to models trained on single distributions.
    \item \textbf{Extrapolation limits \& scaling:} We provide clear evidence that transformers fundamentally struggle with extrapolation outside their syntactic training support. While scaling compute yields consistent improvements, these gains remain strictly log-linear, indicating the necessity for advanced search techniques to improve out-of-support generalization.
\end{itemize}

\noindent \textbf{Reproducibility}: To facilitate future research on transformer generalization, all code, datasets, and the exact model architecture used for this study will be open-sourced. An anonymized repository containing the dataset generators, training splits, and evaluation scripts is provided with the supplementary material.

\section{Related Work}
\label{sec:related_work}
While recent literature often uses the terms ``code generation'' and ``program synthesis'' interchangeably, we ground our study in classic specification-driven program synthesis (specifically, programming by example (PBE)). By isolating this core task, we can rigorously evaluate the foundational generalization limits of modern sequence-to-sequence models before integrating them into systems with additional components.

\textbf{Transformers for code generation.}
The generation of source code has become one of the most prominent applications of transformer models. Foundational progress was made by~\citet{chen2021evaluating}, who introduced \textit{Codex}, the first GPT-based model specialized for programming, alongside the \textit{HumanEval} benchmark and the \textit{pass@k} metric. In parallel, \citet{wang2021codet5} presented \textit{CodeT5}, a multitask encoder–decoder architecture. Since then, large language models (LLMs) for code have scaled substantially in both model size~\cite{nijkamp2022codegen, roziere2023code, li2023starcoder, zhu2024deepseek} and training data volume~\cite{kocetkov2022stack, lozhkov2024starcoder, guo2024deepseek}. 
Crucially, state-of-the-art systems increasingly rely on combining neural generation with advanced search and evolutionary strategies to overcome the limits of raw decoding. \textit{AlphaCode}~\cite{li2022competition} generates massive candidate pools, clusters them by behavioral similarity, and filters them using test cases. More recently, \textit{AlphaEvolve}~\cite{novikov2025alphaevolve} adopts an explicit evolutionary feedback loop where an LLM proposes program mutations, descendants are automatically evaluated, and selection mechanisms balance exploration and exploitation. Modern agentic coding systems~\cite{guo2025deepseek, team2024codegemma, hui2024qwen2} further integrate parallel search across solution candidates, underscoring that generation alone is often insufficient without an underlying search topology.

\textbf{Evaluating transformers for code generation.}
The initial standard for evaluating code generation was \textit{HumanEval}~\cite{chen2021evaluating}, followed by \textit{MBPP}~\cite{austin2021program} for standard-library reliance, and \textit{APPS}~\cite{hendrycks2021measuring} for competitive programming. As models evolved, benchmarks shifted toward complex repository-level and agentic tasks, such as \textit{SWE-bench}~\cite{jimenez2023swe, aleithan2024swe} and \textit{RepoBench}~\cite{liu2023repobench}, where models must navigate full codebases to resolve GitHub issues. 
However, a pervasive limitation across these benchmarks is data contamination: test examples or highly similar instances often inadvertently appear in massive pre-training corpora. To mitigate this, live benchmarks like \textit{LiveCodeBench}~\cite{jain2024livecodebench} and \textit{SWE-Bench-Live}~\cite{zhang2025swe} evaluate models exclusively on problems published after their training knowledge cutoff. Nonetheless, systematic assessment of out-of-distribution generalization remains intractable on these benchmarks because proprietary pre-training datasets obscure the true geometric distance between training and test instances. To solve this, we generate a strictly controlled dataset to study model performance on known training and test distributions.

\textbf{Generalization in transformers.}
Transformer generalization has historically been studied through the lens of \textit{length generalization}---the ability to process sequences longer than those seen during training~\cite{zhou2402transformers}. Empirical results consistently show that models fail to generalize beyond their training distribution boundaries on algorithmic tasks~\cite{fan2024looped, xu_etal_2025_principled}. \citet{zhou2023algorithms} provided theoretical grounding for this via the RASP-L framework, proving transformers only generalize to longer sequences if the optimal, length-independent algorithm requires less representational capacity than a position-dependent shortcut that merely overfits the training data.
In the specific context of program synthesis, \citet{cooper2024generalizability} demonstrated that current positional encodings fail to generalize to unseen code lengths, arguing that reliable performance strictly requires comprehensive training coverage. Furthermore, \citet{zhang2025memorize} found that trivial modifications to standard coding problems (e.g., renaming variables or altering contexts without changing logic) cause catastrophic performance drops, strongly suggesting that LLM generalization is largely just fuzzy retrieval of memorized training instances. Further reinforcing this, \citet{ouellette2025out} showed that on the ARC-AGI benchmark~\cite{chollet2019measure}, LLMs excel at pattern matching but fundamentally struggle to invent novel compositional logic. 

These findings underscore the need for a systematic, metric-driven evaluation of transformer generalization in program synthesis. Our work directly addresses this gap by defining formal probability densities and supports over synthetic program manifolds, allowing us to explicitly measure performance as models are forced beyond their training boundaries.

\section{Methodology}
\label{sec:methodology}
To systematically measure the generalization boundaries of sequence-to-sequence models, we must first construct a rigorously controlled environment. We define a context-free grammar for arithmetic programs and establish two continuous embedding manifolds---semantic and syntactic. This dual-manifold approach allows us to formally quantify the exact behavioral and structural distances between any two programs in the search space.

\begin{figure}[t]
    \centering
    \includegraphics[width=0.99\columnwidth]{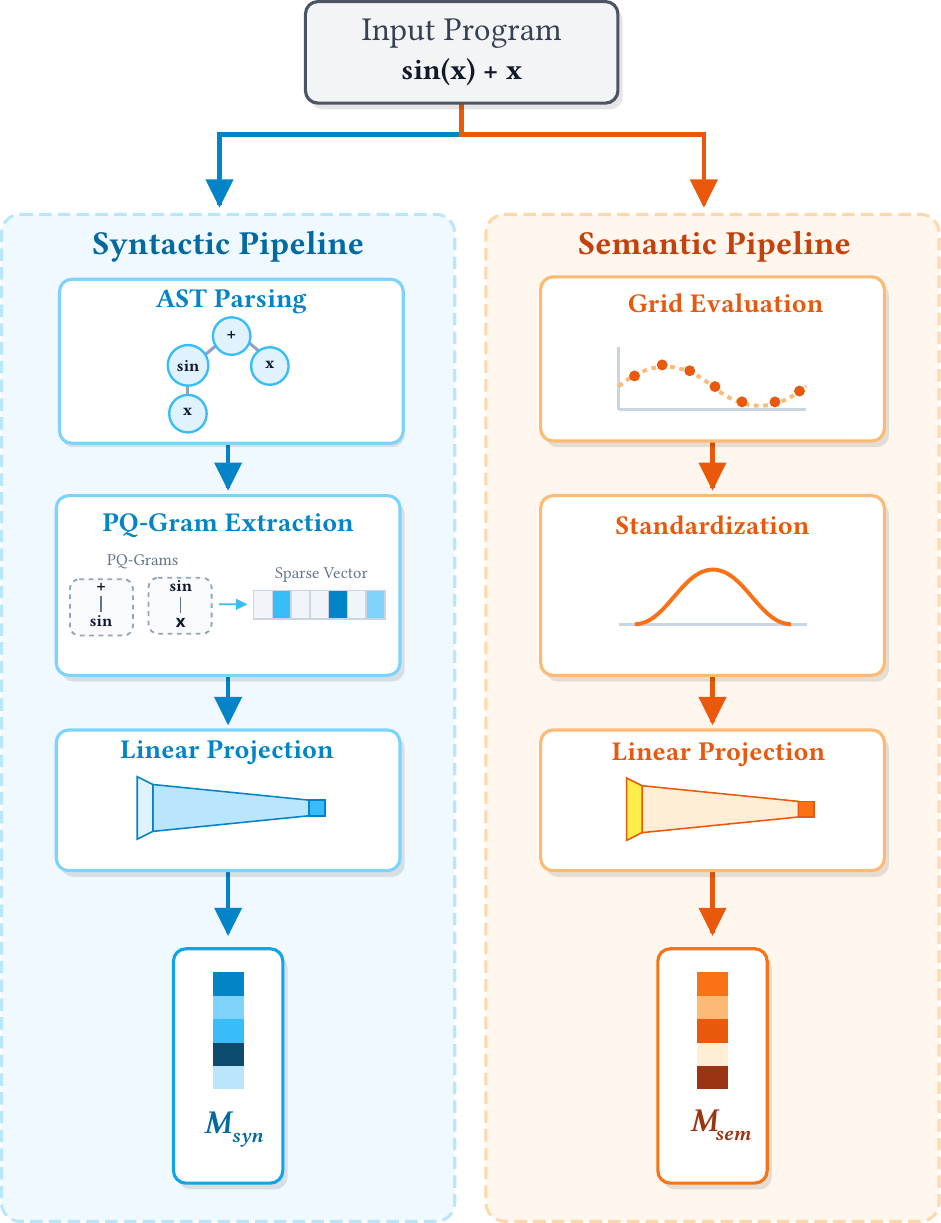}
    \caption{\textbf{Dual manifold projection pipeline.} To formally quantify out-of-distribution distances, arithmetic programs are embedded into two continuous spaces. In the \textbf{syntactic pipeline}, code is parsed into an AST, represented as a sparse vector of PQ-Grams, and condensed using a linear projection (SVD). In the \textbf{semantic pipeline}, the program's functional behavior is evaluated over a fixed input grid, z-score standardized, and also reduced via a linear projection (PCA).}
\label{fig:methodology}
\end{figure}

\subsection{Arithmetic Programs}
\label{subsec:arithmetic_programs}
To isolate the generalization problem from the noise of high-level programming languages, we focus on the domain of univariate (1D input to 1D output) scalar arithmetic. Arithmetic programs have the advantage of a simple syntax, and their semantic behavior can be intuitively understood through their graphical representation. 

\textbf{The domain-specific language (DSL).}
We define a context-free grammar (CFG) $\mathcal{G} = (V, \Sigma, R, S)$ that generates arithmetic expressions acting on an input scalar $x$. The components of the grammar are formally defined as follows:
\begin{itemize}
    \item $V = \{S, E\}$ is the set of non-terminal symbols.
    \item $\Sigma = \{x, +, -, \times, /, \sin, \cos, \exp, \log, \sqrt{\cdot}\}$ is the set of terminal symbols.
    \item $S \in V$ is the designated start symbol.
    \item $R$ is the set of production rules, defined as:
\end{itemize}
\begin{align}
    S &\to E \\
    E &\to E \oplus E \mid \mathcal{F}(E) \mid x \\
    \oplus &\in \{+, -, \times, /\} \\
    \mathcal{F} &\in \{\sin, \cos, \exp, \log, \sqrt{\cdot}\}
\end{align}
Here, $x \in \Sigma$ represents the single input variable evaluated by the program. To prevent infinite generation and bound the search space, we enforce a strict recursion depth limit, allowing a maximum of $L=6$ operator nodes in the abstract syntax tree (AST) of any valid program.

\textbf{Enumeration and reduction.}
We perform a brute-force enumeration of all valid programs in $\mathcal{G}$ up to length $L$, yielding a raw set of programs $\mathcal{P}_{raw}$. A critical challenge in evaluating program synthesis is \textit{semantic ambiguity}---the existence of multiple distinct source codes that compute the exact same mathematical function (e.g., $x+x$ vs. $2 \times x$). To measure true functional generalization rather than the memorization of synonymous token sequences, we must collapse these redundancies.

We define an observational equivalence relation $\sim$ over a discrete, bounded evaluation domain $\mathcal{D}$. Two programs $p_i, p_j \in \mathcal{P}_{raw}$ are considered semantically equivalent if their outputs are numerically identical (up to \texttt{float32} precision) for all inputs:
\begin{equation}
    p_i \sim p_j \iff \forall x \in \mathcal{D}, \; p_i(x) = p_j(x)
\end{equation}
We partition $\mathcal{P}_{raw}$ into equivalence classes $\mathcal{P}_{raw} / \sim$. For each class, we collect all unique, syntactically distinct programs as valid representatives. Our final program universe $\mathcal{U}$ is the set of these unique, syntactic representatives. This guarantees that the downstream dataset covers a controlled maximum of syntactically diverse programs that collectively maximize semantic coverage.

\subsection{The Semantic Manifold ($\mathcal{M}_{sem}$)}

The semantic manifold captures the underlying \textit{behavior} of the programs. To embed a program's behavior into a continuous vector space, we utilize its discretized function plot as a direct representation of its execution semantics (see Figure~\ref{fig:methodology}).

\textbf{Embedding process.}
We define a canonical evaluation interval $I = [a, b]$ and sample $N$ equidistant points within this interval to form an input vector $\mathbf{x}_{eval} \in \mathbb{R}^N$. For any program $p \in \mathcal{U}$, the raw semantic embedding is its output $\mathbf{y} = p(\mathbf{x}_{eval})$.

To ensure our similarity measure captures the \textit{shape} of the function (e.g., linear vs. quadratic growth) rather than merely the magnitude of the output values, we apply z-score standardization to the output vector:
\begin{equation}
    \mathbf{z} = \frac{\mathbf{y} - \mu_{\mathbf{y}}}{\sigma_{\mathbf{y}}}
\end{equation}
where $\mu_{\mathbf{y}}$ and $\sigma_{\mathbf{y}}$ are the scalar mean and standard deviation of vector $\mathbf{y}$. To construct a compact and computationally tractable manifold, we perform principal component analysis (PCA)~\cite{pearson1901liii} over the aggregate dataset of standardized outputs. This projects the $N$-dimensional output vectors down to a dense $d$-dimensional final embedding $\phi_{sem}(p) \in \mathbb{R}^d$ ($d \ll N$; we empirically set $d=32$).

\textbf{Metric definition.}
We define the semantic distance between two programs as the Euclidean ($L_2$) distance between their PCA-projected embeddings, forming the metric space $\mathcal{M}_{sem} = (\mathcal{U}, d_{sem})$:
\begin{equation}
    d_{sem}(p_i, p_j) = \|\phi_{sem}(p_i) - \phi_{sem}(p_j)\|_2
\end{equation}
In this space, $p_i$ and $p_j$ are close if their standardized output behaviors are highly correlated. This naturally clusters programs by algorithmic similarity (e.g., all linear functions cluster together, distinctly separated from cubic functions), regardless of their underlying syntactic implementation.

\subsection{The Syntactic Manifold ($\mathcal{M}_{syn}$)}

Conversely, the syntactic manifold captures the \textit{structure} of the program, strictly independent of its output. To embed programs based on their AST, we utilize PQ-Grams~\cite{augsten2008pq} combined with feature hashing~\cite{weinberger2009feature}.

\textbf{Embedding process.}
First, we decompose the AST of a program $p$ into a multiset of PQ-Grams. A PQ-Gram is a localized subtree composed of a stem, effectively capturing local structural dependencies and hierarchical relationships within the code. We map this multiset into a high-dimensional sparse vector using feature hashing. To project this sparse representation into a dense, numerical vector, we employ a learned linear projection. Given the scale of the matrices involved, we use truncated singular value decomposition (SVD)~\cite{deerwester1990indexing} to obtain a computationally efficient approximation, yielding a dense $d$-dimensional embedding $\mathbf{v}_{svd}$ (aligned with the semantic manifold, we set $d=32$). Finally, to facilitate distance calculations, we apply $L_2$ normalization:
\begin{equation}
    \phi_{syn}(p) = \frac{\mathbf{v}_{svd}}{\|\mathbf{v}_{svd}\|_2}
\end{equation}

\textbf{Metric definition.}
We define the syntactic distance between two programs as the Euclidean ($L_2$) distance between their normalized embeddings, establishing the metric space $\mathcal{M}_{syn} = (\mathcal{U}, d_{syn})$:
\begin{equation}
    d_{syn}(p_i, p_j) = \|\phi_{syn}(p_i) - \phi_{syn}(p_j)\|_2
\end{equation}
In this space, programs are close if they share similar tree structures. This naturally clusters programs by structural similarity (e.g., all ASTs that contain the same syntactic pattern), regardless of their output. 

\section{Experiments}
\label{sec:experiments}
\begin{figure*}[t]
    \centering
     \includegraphics[width=0.99\textwidth]{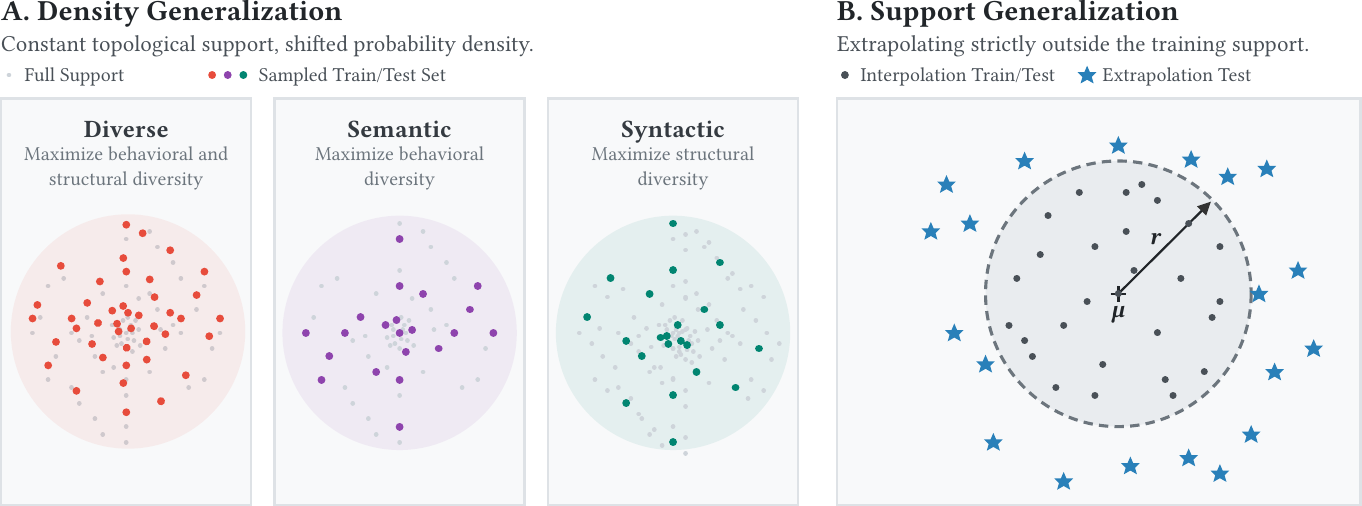}
    \caption{\textbf{Experimental splits in the embedding manifolds.} \textbf{(A) Density generalization:} Isolates the effect of probability mass shifts within a constant support. We compare Diverse sampled data distributions against inverse-density sampling strategies (Semantic and Syntactic). \textbf{(B) Support generalization:} Measures pure extrapolation capabilities by strictly partitioning the space. Models are trained on an interpolation regime (inside radius $r$, gray area) and evaluated on a strictly disjoint out-of-distribution extrapolation regime (outside radius $r$).}
    \label{fig:experiments}
\end{figure*}

Standard evaluation in program synthesis relies heavily on static test splits where the geometric distance to the models' training distribution is completely unknown. To rigorously evaluate generalization bounds, we leverage the semantic and syntactic metric spaces defined in Section~\ref{sec:methodology} to construct a systematic evaluation suite, as conceptually illustrated in Figure~\ref{fig:experiments}.

\subsection{Experimental Design}
\label{subsec:experimental_design}
We formally define a learning task as training on instances sampled from a discrete probability distribution $\mathcal{D}_{train}$ over the universe $\mathcal{U}$, and evaluating on $\mathcal{D}_{test}$. The mass at a point $x\in\mathcal{M}_\star$ (where $\star \in \{ sem,syn\}$) is $\mathcal{D}(\phi_\star^{-1}(x))$, and its support is $\text{supp}_\star (\mathcal{D}) = \{ x\in\mathcal{M} \::\: \mathcal{D}(\phi_\star^{-1}(x)) > 0\}$. We design two experimental protocols to systematically manipulate this distribution.

\subsubsection{Experiment I: Density Generalization}
This experiment isolates the effect of probability mass shifts while holding the support constant ($\text{supp}(\mathcal{D}_{train}) = \text{supp}(\mathcal{D}_{test}) = \mathcal{U}$). It tests whether a model is robust to novel sampling frequencies over known structures or behaviors. 

To approximate uniform sampling over our non-uniform embedding spaces, we apply inverse density sampling~\cite{loftsgaarden1965nonparametric}. The density around a program is estimated by the distance to its $k$-nearest neighbors. Let $d_k(p)$ be the mean distance to the $k$-nearest neighbors of $p$ ($k=5$). The sampling probability is defined proportionally to the inverse density:
\begin{equation}
    P_{sample}(p) \propto \frac{1}{\text{density}(p)} \approx d_k(p)
\end{equation}
Using this technique, we generate three distinct 80/20 (train/test) data splits:
\begin{itemize}
    \item \textbf{Syntactic:} Programs are sampled uniformly over the syntactic embedding space $\mathcal{M}_{syn}$, maximizing the variance of program \textit{structures}.
    \item \textbf{Semantic:} Programs are sampled uniformly over the continuous semantic embedding space $\mathcal{M}_{sem}$. This maximizes the variance of functional \textit{behaviors} seen during training.
    \item \textbf{Diverse:} Programs are sampled uniformly over the set of functional equivalency classes $\mathcal{P}_{raw} / \sim$ containing unique syntactic programs, maximizing \textit{structural} and \textit{behavioral} diversity.
\end{itemize}
We train a separate sequence-to-sequence model on each training split and perform cross-evaluation on all test splits to measure out-of-distribution density generalization.

\subsubsection{Experiment II: Support Generalization}
This experiment tests pure extrapolation---the ability to synthesize novel structures and behaviors outside the support of the training data ($\text{supp}_\star (\mathcal{D}_{train}) \cap \text{supp}_\star (\mathcal{D}_{test}) = \emptyset$).

We utilize a geometric partitioning strategy within the manifolds. Let $\mu$ be the centroid (median vector) of the embedding space and $r$ be a dynamically selected radius threshold. We define two mutually exclusive regions:
\begin{align}
    \mathcal{S}_{in} &= \{p \in \mathcal{U} \mid d(p, \mu) \le r \} \\
    \mathcal{S}_{out} &= \{p \in \mathcal{U} \mid d(p, \mu) > r \}
\end{align}
To ensure controlled sample sizes between regimes, the radius $r$ is chosen such that $80\%$ of the respective set $\mathcal{M}_{sem}$ or $\mathcal{M}_{syn}$ lies within $r$ of $\mu$, and $20\%$ lies outside, thereby partitioning the universe into two mutually exclusive regions. To strictly measure support generalization without density bias, we uniformly sample programs from within these regions. For both $\mathcal{M}_{sem}$ and $\mathcal{M}_{syn}$, we define an \textbf{interpolation} regime (trained and tested on disjoint subsets within $\mathcal{S}_{in}$) and an \textbf{extrapolation} regime (trained on $\mathcal{S}_{in}$, tested exclusively on $\mathcal{S}_{out}$).

\subsection{Experimental Setup}
\label{subsec:experimental_setup}
Having established the theoretical framework, we describe the experimental setup for empirical validation. We first define the task and dataset generation procedure, then present the model architecture and finally outline the training protocol and evaluation criteria.

\textbf{Task formulation and dataset generation.}
We formulate the task as programming by example. The specification $\mathcal{S}$ consists of $N=1{,}000$ input-output pairs derived from a ground-truth program $p^*$. 
To construct our dataset, we exhaustively enumerate all arithmetic programs containing up to $L = 6$ operators. To ensure valid functional semantics, we discard any program that evaluates exclusively to invalid numerical states (e.g., \texttt{NaN} or $\pm\infty$) across the evaluation domain $[-10,10]$. Concretely, we exclude programs for which it is not possible to obtain 1{,}000 distinct valid input--output pairs after $k$ sampling attempts; empirically, we set $k = 1,000,000$. Furthermore, we perform strict structural deduplication such that each valid, executable AST exists exactly once. This procedure yields a rigorously controlled program universe, $\mathcal{U}$, comprising approximately $7$ million unique programs. We partition this set into an 80/20 train-test split. From these splits, we construct our \textit{Semantic}, \textit{Syntactic}, and \textit{Diverse} train and test sets using the sampling strategies defined in Section~\ref{subsec:experimental_design}. The training sets contain $1,000,000$ programs, while the test sets contain $1,000$ programs.

\textbf{Model architecture.}
To process numeric data effectively without introducing tokenization artifacts, we employ a multimodal transformer architecture inspired by LLaVA~\cite{liu2023visual}. Specifically, rather than tokenizing numbers as strings, each numeric pair $(x, y)$ is processed by a lightweight multi-layer perceptron (MLP) projector, which maps the continuous values directly into the input embedding space of an autoregressive transformer decoder. We explicitly utilize this continuous projection rather than a standard byte-pair encoding (BPE) to keep the vocabulary small as in~\cite{biggio2021neural, vastl2024symformer, lalande2023transformer}. The target output is the character-level source code of $p^*$. We evaluate models scaling from $2$ to $8$ layers, utilizing $2$ to $16$ attention heads and a hidden dimension of $64$ to $384$ (further details in the supplementary material).

\textbf{Training details and evaluation.}
Models are trained on a single NVIDIA RTX A6000 GPU using the AdamW optimizer (learning rate $1e^{-4}$, linear warmup). Depending on the layer scale, a single full training run required between $2$ to $24$ hours. We utilize a standard cross-entropy loss on the target code tokens, a batch size of $32$, and early stopping based on validation loss to prevent severe overfitting. For evaluation, we report the \textbf{pass@k} metric ($k \in \{1, 5, 10\}$), as described in~\cite{chen2021evaluating}. A generated program $\hat{p}$ is deemed correct if and only if its output functionally matches $p^*$ across all $N$ inputs. All reported results are averaged across $5$ independent runs, with tables denoting the mean performance $\pm$ one standard deviation.

\subsection{Results and Analysis}
\label{subsec:results}
We structure our analysis around four primary research questions evaluating the limits of transformer generalization.

\paragraph{RQ1: Density generalization.}
\textit{To what extent does model performance degrade when moving out of the training distribution?}

Table~\ref{tab:density_results} presents the cross-evaluation of models trained and tested across the Syntactic, Semantic, and Diverse data splits. 

\begin{table}[t]
\centering
\small
\caption{\textbf{Density generalization (\textit{pass@k}).} Cross-evaluation of models trained on specific distributions. Rows indicate the training distribution, and columns indicate the test split. The \textbf{Diverse} training strategy consistently demonstrates the highest robustness against out-of-distribution probability mass shifts. Best performance per column is shown bold.}
\label{tab:density_results}
\begin{tabular}{llccc}
\toprule
\multirow{2}{*}{\textbf{Train $\downarrow$}} & \multirow{2}{*}{\textbf{Metric}} & \multicolumn{3}{c}{\textbf{Test Distribution $\rightarrow$}} \\
\cmidrule(lr){3-5}
& & \textbf{Diverse} & \textbf{Semantic} & \textbf{Syntactic} \\
\midrule
\multirow{3}{*}{\textbf{Syntactic}}
  & pass@1  & $0.155 \pm 0.002$ & $0.132 \pm 0.002$ & $0.184 \pm 0.002$ \\
  & pass@5  & $0.344 \pm 0.005$ & $0.291 \pm 0.003$ & $0.398 \pm 0.004$ \\
  & pass@10 & $0.415 \pm 0.007$ & $0.357 \pm 0.004$ & $0.480 \pm 0.005$ \\
\midrule
\multirow{3}{*}{\textbf{Semantic}}
  & pass@1  & $0.099 \pm 0.002$ & $\mathbf{0.305 \pm 0.005}$ & $0.106 \pm 0.001$ \\
  & pass@5  & $0.230 \pm 0.005$ & $\mathbf{0.501 \pm 0.008}$ & $0.237 \pm 0.005$ \\
  & pass@10 & $0.294 \pm 0.007$ & $\mathbf{0.549 \pm 0.009}$ & $0.297 \pm 0.007$ \\
\midrule
\multirow{3}{*}{\textbf{Diverse}}
  & pass@1  & $\mathbf{0.193 \pm 0.005}$ & $0.189 \pm 0.002$ & $\mathbf{0.195 \pm 0.002}$ \\
  & pass@5  & $\mathbf{0.398 \pm 0.008}$ & $0.378 \pm 0.006$ & $\mathbf{0.411 \pm 0.003}$ \\
  & pass@10 & $\mathbf{0.471 \pm 0.008}$ & $0.445 \pm 0.007$ & $\mathbf{0.490 \pm 0.003}$ \\
\bottomrule
\end{tabular}
\end{table}

We observe three distinct behavioral trends regarding distribution shifts. 
First, models trained on the \textbf{Diverse} distribution exhibit the highest robustness against out-of-distribution density shifts.
Their \textit{pass@1} performance remains remarkably stable ($\sim 19\%$) regardless of whether the test data is drawn from the in-distribution (Diverse) or out-of-distribution (Semantic/Syntactic) splits. Second, optimizing for semantic diversity during training yields the highest absolute in-distribution test performance: the \textbf{Semantic} model achieves an impressive $30.5\%$ \textit{pass@1} on Semantic test data. However, this comes at the cost of severe degradation (dropping to $\sim 10\%$) when evaluated out-of-distribution on Diverse or Syntactic test sets. The Syntactic model demonstrates a middle ground, experiencing moderate degradation when moving OOD. Finally, we note that overall density generalization remains a heavily bounded challenge; even in the best-case scenarios, the models successfully generalize to at most 10\% to 30\% of the unseen novel problem instances in a pass@1 decoding setting.

\paragraph{RQ2: Support generalization.}
\textit{Can the model extrapolate to novel structures and functional behaviors outside the support of its training data?}

Table~\ref{tab:support_results} contrasts model performance on interpolation ($\mathcal{S}_{in} \to \mathcal{S}_{in}$) versus strict extrapolation ($\mathcal{S}_{in} \to \mathcal{S}_{out}$).
\begin{table}[t]
\centering
\small
\caption{\textbf{Support generalization (\textit{pass@k}).} Evaluating the interpolation vs. extrapolation gap across five independent runs. While evaluating on out-of-support functional behaviors (semantic manifold) incurs virtually no relative penalty, extrapolating to out-of-support syntactic structures causes significant performance degradation.}
\label{tab:support_results}
\begin{tabular}{llccc}
\toprule
\multirow{2}{*}{\textbf{Manifold}} & \multirow{2}{*}{\textbf{Metric}} & \multicolumn{2}{c}{\textbf{Generalization Regime}} & \multirow{2}{*}{\textbf{Gap}} \\
\cmidrule(lr){3-4}
& & \textbf{Interpolation} & \textbf{Extrapolation} & \\
\midrule
\multirow{3}{*}{\textbf{Semantic}}
  & pass@1  & $0.131 \pm 0.003$ & $0.131 \pm 0.001$ & $\mathbf{-0.179\%}$ \\
  & pass@5  & $0.290 \pm 0.006$ & $0.300 \pm 0.002$ & $\mathbf{+3.338\%}$ \\
  & pass@10 & $0.359 \pm 0.007$ & $0.366 \pm 0.001$ & $\mathbf{+2.083\%}$ \\
\midrule
\multirow{3}{*}{\textbf{Syntactic}}
  & pass@1  & $0.188 \pm 0.003$ & $0.130 \pm 0.004$ & $\mathbf{-30.468\%}$ \\
  & pass@5  & $0.360 \pm 0.005$ & $0.259 \pm 0.006$ & $\mathbf{-27.967\%}$ \\
  & pass@10 & $0.416 \pm 0.006$ & $0.305 \pm 0.007$ & $\mathbf{-26.592\%}$ \\
\bottomrule
\end{tabular}
\end{table}
We observe starkly contrasting behaviors in support generalization depending on the embedding manifold. Models trained on the \textbf{Semantic} manifold exhibit low-to-moderate absolute performance within their interpolation regime ($13.1\%$ \textit{pass@1}). However, surprisingly, this performance remains entirely stable when evaluated on pure extrapolation test data ($13.1\%$ \textit{pass@1}), which consists of functional behaviors strictly outside the bounds seen during training. In sharp contrast, models trained within the \textbf{Syntactic} support suffer a massive degradation when forced to extrapolate. While they achieve a higher interpolation baseline ($18.8\%$ \textit{pass@1}), their performance drops considerably by over $30\%$ when generating syntactically novel structures outside their pre-training regime. To put this into context, absolute performance remains relatively low across both manifolds ($13.1\%$ vs. $18.8\%$ \textit{pass@1}). Nevertheless, the exact matching of interpolation and extrapolation performance on the \textbf{Semantic} manifold strongly indicates that the model synthesizes OOD semantic behavior with the exact same proficiency as for in-distribution behavior, provided the underlying syntax remains familiar.

\paragraph{RQ3: Scaling behavior.}
\textit{How is generalization performance influenced by model scale?}

\begin{figure}[t]
    \centering
    \includegraphics[width=\columnwidth]{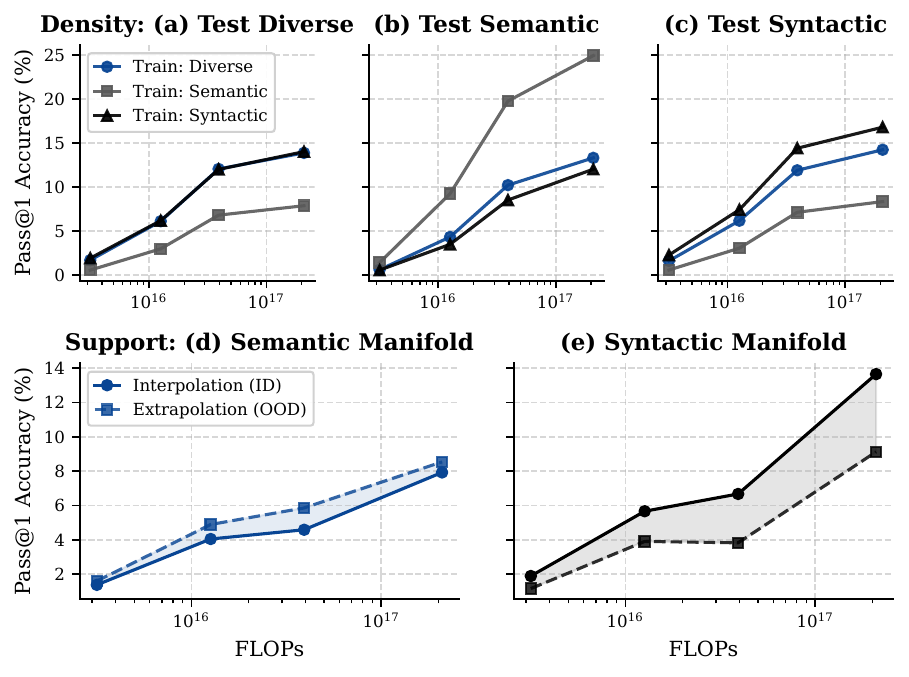}
    \caption{\textbf{Scaling laws for generalization.} Performance (\textit{pass@1}) as a function of compute (FLOPs). \textbf{(a--c) Density Generalization:} All models exhibit strictly log-linear improvements regardless of the sampling strategy. \textbf{(d--e) Support Generalization:} Semantic extrapolation (d) scales identically to interpolation, whereas structural extrapolation (e) suffers a persistent OOD penalty that cannot be easily overcome by simply scaling up compute.}
    \label{fig:scaling_experiments}
\end{figure}
Figure~\ref{fig:scaling_experiments} tracks density generalization (top row) and support generalization (bottom row) against invested compute (FLOPs) by scaling the transformer layers. 
Increasing model scale consistently improves absolute generalization performance across all experimental setups.
However, looking at the density generalization, models trained on datasets emphasizing diversity scale more efficiently and robustly across domains. Conversely, the \textbf{Semantic}-trained model reaches significantly higher peak in-domain performance (Figure~\ref{fig:scaling_experiments}b) but suffers severe scaling penalties when tested out-of-distribution (Figures~\ref{fig:scaling_experiments}a and \ref{fig:scaling_experiments}c). 

Crucially, regardless of the sampling strategy or generalization regime, performance scales strictly \textit{log-linearly}. This indicates a fundamental architectural bottleneck in discovering unseen OOD solutions---exponentially more compute is required for linear gains. Furthermore, focusing on support generalization, Figure~\ref{fig:scaling_experiments}d confirms that semantic interpolation and extrapolation remain tightly coupled across all model scales. In contrast, Figure~\ref{fig:scaling_experiments}e reveals a persistent and widening gap (shaded area) between interpolation and extrapolation on the syntactic manifold, confirming that the structural OOD penalty cannot be easily overcome by simply scaling up FLOPs.

\paragraph{RQ4: Qualitative analysis.}
\textit{What geometric conditions dictate whether a model successfully generalizes to a novel instance?}

\begin{figure}[t]
    \centering
    \includegraphics[width=\columnwidth]{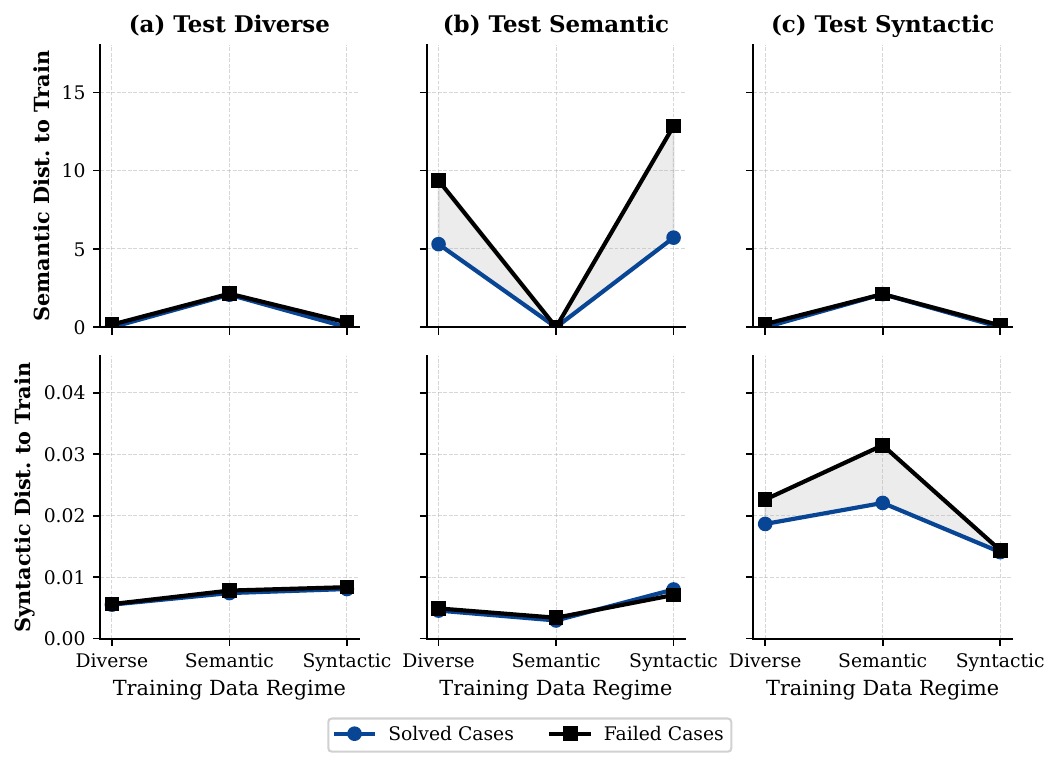}
    \caption{\textbf{Distance to nearest training neighbor.} Comparing distance of solved (blue) vs. failed (black) test instances across semantic (top) and syntactic (bottom) manifolds. When evaluated out-of-distribution (panels b and c), models successfully solve instances that are significantly closer to their pre-training support, highlighting a heavy reliance on local manifold interpolation.}
    \label{fig:qualitative_experiments}
\end{figure}
 
To understand the mechanics of generalization, Figure~\ref{fig:qualitative_experiments} maps the semantic and syntactic distances from solved (blue) and unsolved (black) test cases to their nearest training neighbors. As expected, in-distribution evaluations naturally exhibit the lowest average distance to their respective training sets. Specifically, when evaluating on the \textbf{Diverse} test set (Figure~\ref{fig:qualitative_experiments}a), the overall distance to the training data is uniformly low, with virtually no proximity difference between solved and failed cases. We attribute the superior robustness of the \textbf{Diverse} training strategy directly to this geometric property. By explicitly sampling across both syntactic and semantic variations, this distribution maximizes manifold coverage. This ensures that almost any test instance falls within a densely populated neighborhood, significantly increasing the likelihood that the model has encountered a similar input compared to datasets optimized for a single metric. 

However, a distinctly different pattern emerges under targeted out-of-distribution shifts (Figures~\ref{fig:qualitative_experiments}b and \ref{fig:qualitative_experiments}c). When evaluated outside their primary training distributions, the models exhibit a pronounced gap in nearest-neighbor proximity: the OOD instances they solve are, on average, geometrically closer to the pretraining support than the OOD instances on which they fail. These results suggest that successful transformer generalization may depend, at least in part, on proximity to the training data manifold. The models appear to learn robust representations of their training instances and smoothly interpolate the data manifold within immediate local neighborhoods. Consequently, when forced to evaluate novel instances in sparse regions, the model must construct solutions within these approximated, in-between spaces. Generalization in this context effectively functions as fuzzy behavioral retrieval---interpolating between known training anchors---rather than composition from first principles.

\section{Discussion and Limitations}
\label{sec:discussion}
If transformer-based program synthesis is bounded by local manifold proximity, how should we build the next generation of program synthesis systems? The results of our controlled experiments provide useful insights into the underlying mechanics of generalization in transformer-based program synthesis. By mapping exactly where and why models fail, we can draw clear conclusions for both the curation of training data and the design of future synthesis systems.

\textbf{Diverse sampling balances peak performance and robustness.} 
Our experiments in density generalization (RQ1) demonstrate that sampling training data diversely over unique functional behaviors and distinct abstract syntax trees yields the most robust out-of-distribution performance. While datasets optimized heavily for semantic diversity achieve higher peak performance within distribution, they suffer stronger degradation when forced OOD. 

\textbf{Behavioral vs. structural extrapolation.} 
Interestingly, our support generalization experiments (RQ2) reveal a stark dichotomy in how transformers process novelty. Evaluating models on unseen functional behaviors (semantic extrapolation) incurs virtually no relative penalty compared to interpolation. This suggests that the model's internal representations can successfully synthesize known syntactic operations to match unseen functional forms. We attribute this capability to the model learning robust, abstract representations of functional semantics. When a novel behavior can be analytically decomposed, the transformer successfully maps these intermediate semantic patterns to familiar syntactic structures. While learning this mapping is intrinsically difficult, our empirical scaling laws demonstrate that this ability improves consistently with increased model size.
Crucially, however, this ability is strictly bounded by the model's structural vocabulary. When forced to extrapolate to entirely novel syntactic structures---such as unseen AST combinations or programs longer than those encountered during training---performance drops considerably. Furthermore, we note that the absolute \textit{pass@1} performance on strictly unseen test instances remains fundamentally low across all experimental regimes, which might be a limitation of the model architecture. 

For practitioners, these findings yield a clear directive: achieving robust, out-of-distribution generalization on novel problem instances currently requires scaling to massive models trained on exceptionally large, diversely sampled datasets. Moreover, behavioral extrapolation is significantly more tractable for sequence-to-sequence models than structural extrapolation. If a novel target behavior can be synthesized using familiar structural building blocks, the model is likely to succeed when scaled up enough. Conversely, inventing fundamentally new algorithmic structures remains the primary bottleneck (e.g. synthesizing a solution made up of novel AST component combinations or a solution that is longer than programs seen during training). For the genetic programming community, this presents a highly promising opportunity: it strongly motivates the development of hybrid systems where the neural model acts as a behavioral synthesizer, while evolutionary search operators are explicitly deployed to inject structural novelty and overcome the transformer's syntactic extrapolation deficits.

\textbf{The log-linear bottleneck of generalization.} 
While scaling compute budgets (FLOPs) continuously improves absolute generalization performance (RQ3), this relationship remains strictly log-linear across both interpolation and extrapolation regimes. Consequently, relying purely on model scale to achieve out-of-support synthesis yields severely diminishing returns. Requiring exponentially more compute to achieve linear performance gains on novel instances exposes a fundamental architectural bottleneck in pure autoregressive transformers. This cost-efficiency trade-off contextualizes the massive parameter counts of state-of-the-art large language models currently dominating program synthesis benchmarks. Furthermore, our qualitative analysis (RQ4) hints that a model's success depends on the geometric proximity of the test instance to its nearest training neighbor. This mechanism explains the paradox between the extraordinarily high scores (often exceeding 90\% pass@1 rates) reported on modern coding benchmarks and the modest absolute generalization observed in our controlled extrapolation experiments. When proprietary pre-training datasets reach internet-scale, the geometric distance to almost any benchmark problem shrinks drastically. What appears to be zero-shot algorithmic reasoning is often just dense manifold interpolation.

For practitioners, these dynamics offer clear deployment guidelines. If the objective is exceptional synthesis performance within a known, well-defined target domain, success hinges on efficient dataset curation. Generating a high-quality pre-training corpus that is explicitly sampled to maximize both syntactic and semantic diversity will saturate the local manifold, maximizing in-distribution reliability and providing strong anchor points for near-OOD interpolation. Conversely, if the target domain is open-ended, highly dynamic, or requires solutions structurally disjoint from historical data, static pre-trained models will be fundamentally insufficient. In these strict out-of-distribution scenarios, relying solely on parametric memory is inherently fragile. A promising avenue for future systems might be hybrid paradigms---coupling neural generation with symbolic search, online evaluation, and continuous  learning---to actively synthesize and verify novel solutions on the fly.

\subsection{Implications for Evolutionary Computation}
These findings present a compelling opportunity for the genetic programming (GP) and evolutionary computation communities. If transformers act primarily as powerful behavioral interpolators but struggle with structural extrapolation, they naturally complement the strengths of evolutionary search:
\begin{itemize}[leftmargin=15pt, itemsep=3pt, topsep=3pt]
    \item \textbf{LLMs for local interpolation, GP for global extrapolation:} In hybrid systems, transformers can be utilized as mutation operators to navigate dense, well-known semantic neighborhoods. Concurrently, evolutionary search can be deployed to systematically explore unknown regions where LLMs lack training support. This paradigm has shown success in systems like AlphaEvolve~\cite{novikov2025alphaevolve} and FunSearch~\cite{romera2024mathematical}, which utilize evolutionary search to push LLMs into novel regions of the search space.

    \item \textbf{Continuous learning loop:} The diverse sampling strategy proposed here could be used to train LLMs which generate initial populations for genetic algorithms. The GP environment then systematically expands those initial populations, retains successful individuals, and adds them back to the  training set, expanding the support in an iterative loop.
\end{itemize}

\subsection{Limitations and Future Work}
While our environment provides strict guarantees regarding OOD distances, we acknowledge several limitations in our setup. 

\noindent \textbf{Construct validity:} Our programmatic universe is bound by a maximum AST depth of $L=6$ within a 1D scalar arithmetic DSL. While this ensures exhaustive enumeration and strict observational equivalence, it does not fully capture the complex, multi-modal dependencies of repository-level software engineering tasks. 

\noindent \textbf{External validity:} Scaling the continuous semantic manifold ($\mathcal{M}_{sem}$) to Turing-complete, general-purpose languages (like Python or C++) presents non-trivial challenges, particularly due to the Halting Problem, infinite loop handling, and complex state mutations. Future work must investigate approximate semantic hashing to extend this framework to general-purpose languages.

\noindent \textbf{Reasoning paradigms:} In our current experiments, we evaluated the direct \textit{pass@k} performance of models in a strict PBE scenario. However, state-of-the-art code generation models increasingly rely on Chain-of-Thought (CoT) reasoning, producing intermediate solutions before generating the final program. We did not evaluate these multi-step reasoning scenarios. Extending this fundamental setup to investigate how density and support generalization are influenced by intermediate reasoning steps remains a highly promising direction for future work.

\section{Conclusions}
\label{sec:conclusions}
In this work, we systematically analyzed the generalization boundaries of transformer-based models in program synthesis by rigorously disentangling \textit{density generalization} from \textit{support generalization}. By projecting a controlled arithmetic grammar into continuous syntactic and semantic manifolds, we were able to formally quantify out-of-distribution distances and observe how models behave when forced beyond their training data distributions.
Our findings reveal that density generalization can be vastly improved when training data is sampled diversely across the semantic and syntactic spaces rather than relying on a single criterion (e.g., optimizing exclusively for syntactic diversity or exclusively for semantic diversity). However, evaluating support generalization reveals a notable limitation: while transformers exhibit strong interpolation capabilities within the support of their syntactic training data, their ability to extrapolate to novel syntactic regions drops markedly.
Furthermore, our qualitative and scaling analyses indicate that a model’s success on a novel problem is strongly associated with its proximity to the nearest training instance. While scaling up compute yields consistent generalization improvements, these gains remain strictly log-linear. 
We conclude that maximizing the spread of training data across semantic and syntactic manifolds yields a highly effective, compute-efficient strategy for practitioners training code generation models. 
Ultimately, the log-linear bottleneck of pure neural generalization suggests that scaling alone is insufficient. Building robust, open-ended program synthesis systems may benefit from hybridizing neural models with symbolic search or evolutionary techniques capable of navigating beyond the boundaries of the learned data manifold.


\begin{acks}
This work was supported by the Carl Zeiss Stiftung (“Interactive Inference” project and “CZS Stiftungsprofessuren”). 
\end{acks}

\bibliographystyle{ACM-Reference-Format}
\bibliography{bibliography}



\clearpage


\section*{Supplementary material}

# 1 Details on Arithmetic Programs

## 1.1 Syntax and Semantics

**Formal syntax.** To isolate the generalization problem from the noise of high-level programming languages, we focus on the domain of univariate (1D-input to 1D-output) scalar arithmetic. Arithmetic programs have the advantage of a simple syntax, and their semantic behavior can be intuitively understood through their graphical representation. We define a context-free grammar (CFG) $\mathcal{G} = (V, \Sigma, R, S)$ that generates arithmetic expressions acting on an input scalar $x$. The components of the grammar are formally defined as follows:

- $V = \{S, E\}$ is the set of non-terminal symbols.
- $\Sigma = \{x, +, -, \times, /, \sin, \cos, \exp, \log, \sqrt{\cdot}\}$ is the set of terminal symbols.
- $S \in V$ is the designated start symbol.
- $R$ is the set of production rules, defined as:

$$S \rightarrow E \quad (1)$$
$$E \rightarrow E \oplus E \mid \mathcal{F}(E) \mid x \quad (2)$$
$$\oplus \in \{+, -, \times, /\} \quad (3)$$
$$\mathcal{F} \in \{\sin, \cos, \exp, \log, \sqrt{\cdot}\} \quad (4)$$

Here, $x \in \Sigma$ represents the single input variable evaluated by the program.

**Constants.** Mathematical expressions generally allow for the inclusion of continuous constants. To keep our setup minimal, bounded, and strictly reproducible, we restrict constants to those that emerge naturally from the enumeration process of the grammar. For example, subtracting the input from itself $(x - x)$ resolves to the constant 0, which can then be utilized by subsequent operations within the abstract syntax tree, for instance, $1 = \exp(0) = \exp(x - x)$.

**Formal semantics.** The semantics of an arithmetic program are "what the program does". This behavior is determined by the composition of operators within the AST. An arithmetic program is executed in a bottom-up manner: leaf nodes (the input $x$) are passed into their parent operators, and the partial results are propagated up the tree. By definition, the result evaluated at the root node of the AST is interpreted as the final output of the program.

**Illustrating behavior.** In the example provided in Figure 1, the leaf nodes $x$ are passed into sin and cos operators. The right branch is then passed into an exp operator, while the left partial result is multiplied by $x$. Both branches converge at the root subtraction node, yielding the final output of the program. In contrast to high-level programming languages, the semantic behavior of these arithmetic programs can be readily visualized. Figure 1 illustrates that despite strict bounds on program length, the language covers a diverse and sophisticated range of semantic behaviors across the input domain.

## 1.2 Analysis of the Program Space

Intuitively, due to the vast search space of arithmetic programs, there are many structurally distinct programs that compute the exact same mathematical function $f : \mathbb{R} \rightarrow \mathbb{R}$. Programs that compute the identical function $f$ over the evaluation domain are termed *semantically equivalent*.

**Equivalence classes of arithmetic programs.** Let $f : \mathbb{R}^D \rightarrow \mathbb{R}$ be a function. The equivalence class of $f$ is the following set of arithmetic programs,

$$\{P \mid \forall x \in \mathbb{R}^D, \ P(x) = f(x)\}$$

For instance, the arithmetic programs

$$\sqrt{x \times x} \quad \text{and} \quad x$$

both belong to the equivalence class of the function $f(x) = x$. Categorizing programs into equivalence classes allows us to distill the search space down to a distinct set of unique functional behaviors. Conceptually, it is sufficient to consider only one representative program from each equivalence class. We designate a program as *minimal* for its equivalence class if it possesses the minimal structural length (fewest operators) among all programs in that class. If multiple minimal programs exist for a single class, they are considered *minimal alternatives*.

**Growth behavior of arithmetic program space.** The total number of valid ASTs generated by our grammar grows exponentially with respect to the maximum allowed depth or number of operators $L$. Because each binary operator branches into two sub-trees, the combinatorial explosion of possible operator and leaf arrangements yields a massive number of syntactically valid programs even for small values of $L$.

Specifically, for $D$-dimensional inputs, the number of programs with $L$ operators satisfies the recursion

$$S_D(1) = D\,|I_1| + \binom{D}{2}|I_2|$$

$$S_D(L) = S(L-1)\left((D+L-1)|I_1| + \binom{D+L-1}{2}|I_2|\right),$$

where $|I_1|$ and $|I_2|$ are the numbers of unary and binary instructions in the instruction set $I$. Unrolling this recursion yields the following closed-form expression:

$$S_D(L) = \prod_{j=0}^{L-1} (D+j)\left(|I_1| + |I_2|\frac{D+j-1}{2}\right),$$

which grows exponentially with respect to $L$. Moreover, any arithmetic program with input dimension $D$ has a length of at least $D-1$. To verify that two arithmetic programs $p_1$ and $p_2$ are equivalent, we require that for all inputs $x \in \mathbb{R}^D$:

$$p_1(x) = p_2(x).$$

Note, that in our experiments we consider only the univariate case with $D = 1$. To illustrate the syntactic growth (the number of distinct program structures) and the semantic growth (the number of distinct observed behaviors), we enumerate all arithmetic programs with $D = 1$ up to a maximum length of $L = 4$ operators. We then determine the number of equivalence classes by identifying semantically equivalent programs using the equivalence test described above. All semantic equivalency checks are conducted using SymPy [4]. This assigns each program to an equivalence class.

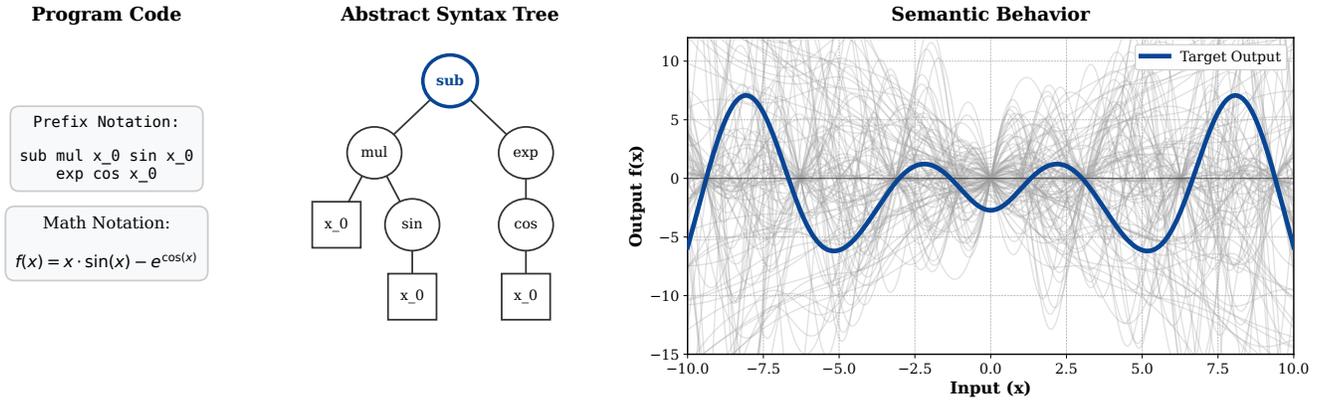

Figure 1: The relationship between a program's syntax and its semantic behavior. On the left, a single program sampled from the arithmetic grammar. On the right, its corresponding function plot, $f(x)$, is highlighted in blue. The faint gray lines illustrate the diverse behaviors of other randomly sampled programs from the same grammar, providing a visual context for the variety of functions the language can express.

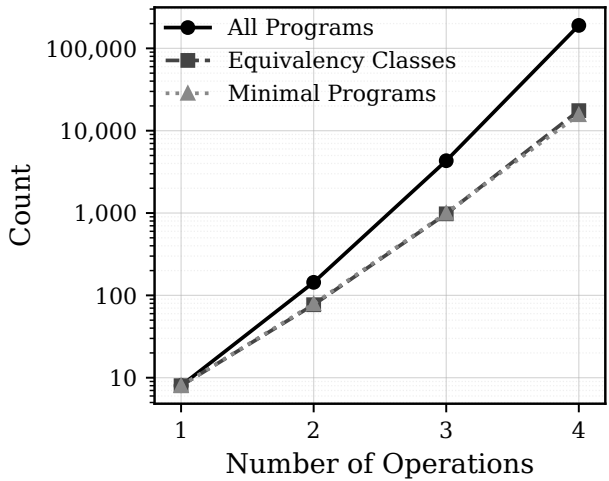

Figure 2: Characterizing the growth behavior of the arithmetic program space. A comparison of syntactic versus semantic growth as a function of program length (number of operations). The total number of syntactically valid programs grows exponentially, far outpacing the growth in the number of unique semantic behaviors (equivalency classes). This demonstrates a high degree of redundancy within the language.

Note, while our main dataset enumerates up to $L = 6$, we restrict this specific symbolic analysis to $L = 4$ due to the more expensive computational overhead of exact symbolic verification at larger scales. Figure 2 illustrates the growth of the total number of distinct structures versus the growth of distinct semantic behaviors (equivalence classes). The figure shows that the number of syntactic structures grows exponentially, whereas the number of distinct behaviors grows up to an order of magnitude more slowly, indicating ambiguities (semantic redundancies) in the language.

**Structural properties of the arithmetic program space.** To demonstrate the rugged and discontinuous nature of the program space, we temporarily define two distance measures: Tree Edit Distance (TED, syntactic) and Normalized Root Mean Squared Error (NRMSE, semantic). *Note: These metrics are used exclusively for the analysis in this section. The evaluation in the main text relies on the continuous PQ-Gram and PCA embeddings ($\mathcal{M}_{syn}$ and $\mathcal{M}_{sem}$).*

- **Tree Edit Distance (TED):** The TED measures the similarity between two syntax trees based on the minimum number of edits (inserts, deletes, updates) required to transform one into the other. Following the algorithm of [6] and assuming a unit cost for each edit operation, we define a similarity score as follows:

$$\text{TED}(T_{\text{true}}, T_{\text{pred}}) = \text{dist}(T_{\text{true}}, T_{\text{pred}}) \quad (5)$$

Here, $\text{dist}(T_{\text{true}}, T_{\text{pred}})$ is the raw minimum edit distance returned by the Zhang-Shasha algorithm, and $|T|$ denotes the number of nodes in a tree. For example, transforming $(x + y) \times z$ to $(x + y) + z$ requires one "update" operation (changing × to +). With 5 nodes in each tree, the TED score would be 1, reflecting their low edit distance.

- **Normalized Root Mean Squared Error (NRMSE):** The Normalized Root Mean Squared Error quantifies the difference between predicted and true continuous values, normalized to ensure the score lies within the range $[0, 1)$. Given the true values $\mathbf{y}_{\text{true}}$ and the predicted values $\mathbf{y}_{\text{pred}}$, it is defined as:

$$\text{NRMSE}(\mathbf{y}_{\text{true}}, \mathbf{y}_{\text{pred}}) = \frac{\sqrt{\frac{1}{n} \sum_{i=1}^{n} \left(y_{\text{true},i} - y_{\text{pred},i}\right)^2}}{1 + \sqrt{\frac{1}{n} \sum_{i=1}^{n} \left(y_{\text{true},i} - y_{\text{pred},i}\right)^2}} \quad (6)$$

Here, the numerator represents the standard Root Mean Squared Error (RMSE), and the denominator normalizes the metric to keep it bounded below 1. A value of 0 indicates a

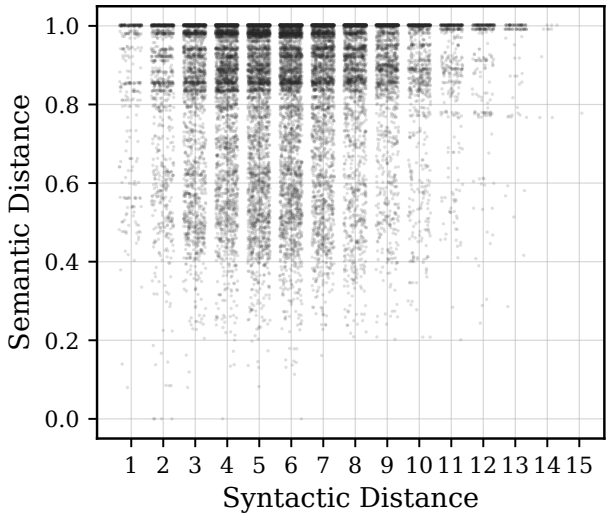

**Figure 3: Analyzing the discontinuity of the program space.** This figure illustrates the relationship between syntactic and semantic distance for pairs of programs. Syntactic distance is measured by the tree edit distance (TED) on the programs' abstract syntax trees, while semantic distance is quantified by the normalized root mean squared error (NRMSE) between their outputs. The plot reveals that even programs with small syntactic differences can have large semantic differences, illustrating the discontinuous and rugged nature of the search space.

perfect prediction ($y_{true} = y_{pred}$), while values approaching 1 indicate larger prediction errors. Note that the NRMSE as defined above is symmetric and therefore satisfies the conditions of a distance measure.

**Discontinuity of the program space.** Given these distance measures, we construct pairwise distance matrices $D^{syn}$ and $D^{sem}$. From our enumeration of programs up to $L = 4$, we select a representative subset via stratified sampling ($n = 1{,}000$) and compute all pairwise distances.

We plot the resulting distance tuples ($d_{syn}, d_{sem}$) for every unordered pair in Figure 3. The horizontal axis corresponds to syntactic tree edit distance, and the vertical axis to semantic NRMSE distance. The scatter plot confirms a critical qualitative observation: numerous pairs of programs exhibit very small syntactic distances (left side of the x-axis) but vastly different semantic distances (near 1 on the y-axis). Conversely, some programs are syntactically dissimilar but compute the exact same function (semantic distance of 0).

## 2 Details on Experiments

We introduced the arithmetic program synthesis task to develop an intuitive understanding of program synthesis in a controlled, low-level setting. The goal of program synthesis is to generate a program that satisfies a given specification. In the arithmetic program synthesis task, the specification is a set of input–output examples produced by the arithmetic program. Specifically, each example is obtained by sampling inputs and recording the outputs corresponding to the mathematical expression computed by the arithmetic program. The objective of a synthesis algorithm is to produce an arithmetic program that is consistent with the provided examples — in other words, a program that passes the examples as test cases. This setup mirrors higher-level benchmarks, such as HumanEval [1], where candidate programs are evaluated based on their ability to pass a series of input–output tests. Success is measured using the standard *pass@k* metric [1], which is the fraction of tasks for which at least one of $k$ generated candidates passes all test cases. In the following sections, we detail the dataset generation process, the evaluated transformer model architecture, and the evaluation metrics used.

---

**Algorithm 1** Arithmetic Program Dataset Generation

**Require:** Max length $L_{max}$, Instructions $I$, Verif. set $X_{val}$, Train sample size $n$
**Ensure:** Dataset $\mathcal{D}$ of unique program examples
1: **Phase 1: Exhaustive Enumeration**
2: $\mathcal{P} \leftarrow$ Generate all valid program DAGs up to length $L_{max}$ using $I$

3: **Phase 2: Equivalence Check & Filtering**
4: $\mathcal{P}_{valid} \leftarrow$ Remove programs failing to yield 1,000 valid outputs in $10^6$ attempts
5: $\mathcal{H} \leftarrow$ Empty Dictionary　　　　　▷ *To store unique behaviors*
6: **for** each program $p \in \mathcal{P}_{valid}$ **do**
7: 　　$h \leftarrow$ Hash($p(X_{val})$)　　　　▷ *Hash the program's outputs*
8: 　　**if** $h \notin \mathcal{H}$ **then**
9: 　　　　$\mathcal{H}[h] \leftarrow \emptyset$　　▷ *Initialize set for this semantic behavior*
10: 　　**end if**
11: 　　**if** $p \notin \mathcal{H}[h]$ **then**
12: 　　　　$\mathcal{H}[h] \leftarrow \mathcal{H}[h] \cup \{p\}$　▷ *Keep all syntactically unique ASTs*
13: 　　**end if**
14: **end for**

15: **Phase 3: Data Formatting**
16: $\mathcal{D} \leftarrow \emptyset$
17: **for** each semantic behavior class $C \in \mathcal{H}$.values() **do**
18: 　　**for** each syntactically unique program $p \in C$ **do**
19: 　　　　$X_{train} \leftarrow$ Sample $n$ random input points
20: 　　　　$Y_{train} \leftarrow p(X_{train})$　　　　▷ *Evaluate program*
21: 　　　　$T \leftarrow$ PreOrderTokenize($p$)　　▷ *e.g., [exp, cos, x]*
22: 　　　　$\mathcal{D} \leftarrow \mathcal{D} \cup \{(X_{train}, Y_{train}, T)\}$
23: 　　**end for**
24: **end for**
25: **return** $\mathcal{D}$

---

### 2.1 Dataset Generation

**Enumeration algorithm.** We employ the exhaustive enumeration algorithm detailed in Algorithm 1 to generate a high-quality dataset of syntactically and semantically unique arithmetic programs:

(1) **Hyperparameters (phase 0).** We determine the input dimension $D = 1$ and a maximum structural length $L_{max} = 6$ operators. This bounds the search space while allowing for a sufficiently complex and rigorously controlled program universe $\mathcal{U}$.

(2) **Enumeration (phase 1).** From the set of unary and binary instructions $I$, we generate all valid operator sequences as skeletons for the arithmetic program's directed acyclic graph (DAG). We then instantiate each DAG skeleton into a list of valid executable programs by generating all valid combinations of input wirings.

(3) **Equivalence check (phase 2).** We generate a validation set $X_{val}$ of $N = 10{,}000$ random input points sampled uniformly from the interval $[-10, 10]$. We infer all generated programs on this validation set. If a program yields NaN or infinite values for more than 95% of the points, it is rejected as mathematically unstable. To filter semantic redundancies, we hash the valid output arrays. In the event of a hash collision (semantic equivalence), we keep the arithmetic program only if it is a unique, unseen syntactic structure, otherwise it is discarded.

(4) **Data formatting (phase 3).** Our deduplication yields a universe $\mathcal{U}$ comprising approximately 7 million syntactically unique arithmetic programs. To construct the final training and test examples, we sample $n = 1{,}000$ input points $X_{train}$ uniformly from $[-10, 10]$, evaluate the program on those points, and collect the resulting input–output pairs. To represent the target source code, we parse each program into its abstract syntax tree and serialize the tree using a pre-order traversal to obtain a token sequence. For example, the mathematical function $e^{\cos(x^2)}$ is tokenized as [exp, cos, square, x].

From this massive universe of 7 million programs, we sample specific subsets of 1,000,000 training programs and 1,000 test programs for our various experimental ablations (Semantic, Syntactic, Diverse) as defined in the main text.

### 2.2 Dataset Split Visualization

In this section, we visualize the distributions of the various data splits to intuitively illustrate how they differ in terms of their spatial composition across the continuous metric spaces.

**Density generalization splits.** Figure 4 visualizes the datasets used for experiment I (density generalization). By naturally sampling unique equivalence classes, the **Diverse** split forms a dense cluster near the origin of both manifolds, representing the most common structural and behavioral patterns. Conversely, the **Semantic** and **Syntactic** splits are constructed using inverse-density sampling. As shown in the local density histograms, this technique effectively flattens the distribution, forcing the training data to sparsely cover the outer boundaries of the manifolds and artificially

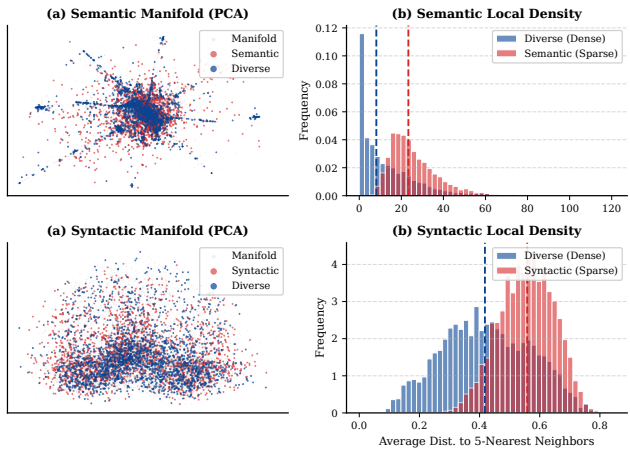

Figure 4: Visualizing density generalization splits. Projection of the Diverse, Semantic, and Syntactic datasets onto the semantic (top) and syntactic (bottom) manifolds. The scatter plots (a) illustrate the spatial coverage of the respective sampling strategies. The histograms (b) display the local density, measured as the average distance to the 5-nearest neighbors. Inverse-density sampling (red) successfully flattens the distribution compared to the naturally center-dense Diverse sampling (blue), forcing the training data to cover sparser, more isolated regions of the manifolds.

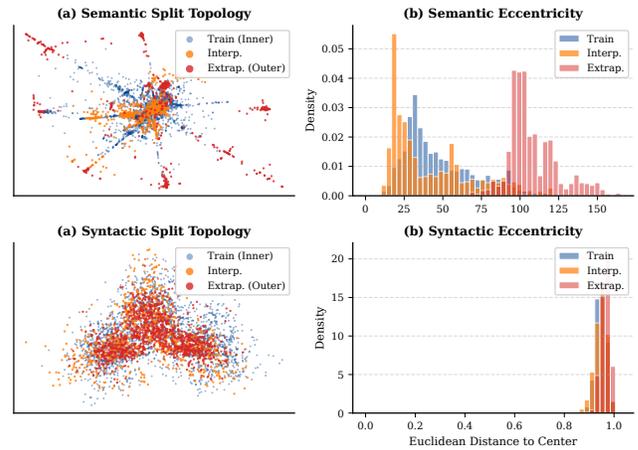

Figure 5: Visualizing support generalization splits. Geometric partitioning of the program space into interpolation and extrapolation regimes. The scatter plots (a) show the distribution of the splits within the semantic (top) and syntactic (bottom) manifolds. The histograms (b) plot the eccentricity (Euclidean distance to the manifold centroid). A strict radius threshold successfully isolates the training and interpolation sets (inner region) from the strictly out-of-distribution extrapolation set (outer region).

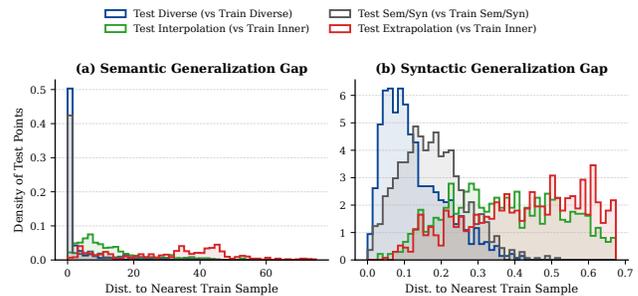

Figure 6: Quantifying out-of-distribution distances. Distribution of test instances based on their exact geometric distance to the nearest training sample in the semantic (a) and syntactic (b) manifolds. The Diverse and Interpolation test sets remain close to the training support. In contrast, the Extrapolation and inverse density sampled (Sem/Syn) test sets exhibit significantly higher distances to the nearest training samples.

shifting the probability mass away from the dense center.

**Support generalization splits.** Figure 5 visualizes the geometric partitioning utilized in experiment II (support generalization). To test pure extrapolation, the dataset is strictly divided based on a radius threshold $r$. The scatter plots confirm that the *interpolation* regime (orange) is strictly confined within the support of the training data (blue). In contrast, the *extrapolation* regime (red) lies entirely outside this boundary, forcing the model to synthesize solutions in regions of the manifold where the training support is strictly zero.

**Quantifying the generalization gap.** To mathematically verify the strictness of our out-of-distribution evaluation, Figure 6 plots the distance from every test instance to its nearest neighbor in the training set. For the **Diverse** and **Interpolation** test sets, the distance to the training data remains close to zero. However, for the **Extrapolation** and uniform density splits, the geometric distance to the nearest training sample shifts significantly higher. This confirms that evaluating on these splits constitutes a rigorous measure of out-of-distribution generalization.

### 2.3 Model Architecture

The transformer architecture in our experiments follows a multi-modal transformer architecture inspired by LLaVA [3] (see Figure 8). To encode the continuous data points $(x, y) \in \mathbb{R} \times \mathbb{R}$, the numerical values are first mapped into the range $[-1, 1]$ using a tanh operator. Subsequently, each 2D coordinate pair is projected into the 384-dimensional embedding space of the language model via a linear projection network. This yields a sequence of $n$ continuous point tokens in the model's context, which are then followed by the <start> and <end> tokens encapsulating the target arithmetic program sequence. All hyperparameters of the model architecture and training process are illustrated in Table 1. From each dataset split, we take back a validation set of 1000 programs. Then we train the model using the standard cross entropy loss until the loss on

Figure 7: Nearest neighbors in the semantic manifold. This figure illustrates a query program's output (thick gray) alongside its three nearest neighbors (N1, N2, N3) in the continuous semantic embedding space. Closeness in this manifold corresponds to highly correlated functional behaviors (Pearson's $r \geq 0.99$). As shown, the embedding successfully captures the overarching shape of the mathematical function, even if the individual programs exhibit minor local phase or amplitude variations.

Table 1: Transformer architecture and training hyperparameters. The top section details the structural parameters varied during the scaling experiments to evaluate generalization as a function of compute. The bottom section outlines the fixed optimization hyperparameters applied uniformly across all model training runs.

| Parameter | Value |
| --- | --- |
| Embedding dimension | 64, 128, 192, 384 |
| Number of decoder layers | 2, 4, 6, 8 |
| Number of attention heads | 2, 4, 8, 16 |
| Batch size | 32 |
| Learning rate | $10^{-4}$ |
| Warmup iterations | 2,000 |
| Weight decay | $10^{-1}$ |
| Adam betas | (0.9, 0.95) |
| Dropout | 0.0 |
| Grad clip | 0.5 |

the validation set is saturated. We train several model ablations with respect to the intended experiment at hand by varying dataset composition and compute budget (FLOPs), as described in the paper.

## 2.4 Evaluation Metrics

We measure model performance in alignment with established program synthesis benchmarks for high-level programming languages (e.g., HumanEval [1]) using the *pass@k* metric. In our programming by example setup, a generated arithmetic program is counted as functionally correct if its output matches the target output across all continuous test points.

**The pass@k metric.** The *pass@k* metric evaluates functional correctness by estimating the probability that at least one of $k$ candidate programs, sampled from a larger generated pool of $n$ programs, will pass all unit tests. It provides an unbiased estimator for this probability, defined formally as:

$$\text{pass@}k = \mathbb{E}_{\text{problems}} \left[ 1 - \frac{\binom{n-c}{k}}{\binom{n}{k}} \right] \quad (7)$$

Here, for a single problem specification, $n$ is the total number of candidate programs generated by the model, $c$ is the number of those candidates that are functionally correct, and $k$ is the number of samples considered ($k \leq n$). The term $\binom{n}{k}$ denotes the standard binomial coefficient. The final score is the average over all problem specifications in the evaluation set.

A score of 1.0 indicates that a correct program is always found within $k$ samples, while a score of 0.0 indicates that a correct program is never found. For instance, a *pass@1* score of 0.5 means the model's single best-guess program is correct for 50% of the problems, whereas a *pass@100* of 0.8 implies an 80% chance of finding at least one correct program among 100 candidates. This metric is particularly relevant for our setup, as it correctly rewards a generated program if it is functionally equivalent to the ground truth (semantic match), even if its underlying sequence of operations differs entirely (syntactic mismatch). In our experiments we use

Figure 8: Multi-modal transformer architecture. Inspired by vision-language models, we bypass standard byte-pair encoding (BPE) for the numerical inputs to prevent tokenization artifacts. Instead, continuous input-output execution pairs are processed by a point encoder and linearly projected directly into the transformer's continuous embedding space. The decoder then autoregressively generates the corresponding sequence of discrete arithmetic program tokens.

$N = 1000$ input-output examples for each test instance. We report the dataset-level *pass@1*, *pass@5*, and *pass@10* rates.

## 3 Additional Experimental Results

In this section, we expand upon the empirical findings presented in the main text by investigating two supplementary research questions. First, we examine the impact of test-time compute (RQ5) to determine whether increasing the sampling budget $k$ at inference time can overcome the models' observed structural extrapolation deficits. Second, we provide a detailed qualitative analysis of the generation failures (RQ6) to characterize the relationship between target program complexity, syntactic validity, and semantic approximation.

*RQ5: Test-time scaling. How is generalization performance influenced by test-time scaling, i.e., the number of sampled candidates $k$?*

**Density generalization.** Recent studies suggest that optimally scaling test-time compute can sometimes surpass the benefits of pre-training scale [5]. We investigate whether this holds true for density and support generalization. Figure 9 expands the density generalization scaling laws across different test-time sampling budgets ($k \in \{1, 5, 10\}$). As expected, increasing the number of sampled candidates reliably improves absolute performance across all training distributions and model scales. However, test-time scaling does not alter the fundamental dynamics of the models. The scaling trajectory remains strictly log-linear regardless of the budget and even seems to show signs of saturation. Furthermore, the robustness advantage of the **Diverse** training strategy persists at higher $k$ values. It consistently avoids the severe out-of-distribution performance drops seen when the **Semantic** and **Syntactic** models are evaluated outside their native distributions. This indicates that the OOD robustness gained from diverse sampling is a deeply ingrained property of the learned representations, not merely an artifact of greedy decoding.

**Support generalization.** Figure 10 evaluates whether test-time scaling can overcome the model's inability to extrapolate to novel structures. The data clearly shows it cannot. On the semantic manifold (top row), interpolation and extrapolation performance remain tightly coupled, increasing in tandem as $k$ increases. On the syntactic manifold (bottom row), however, the large performance gap (shaded area) between interpolation and extrapolation persists across all sampling budgets. In fact, in absolute terms, the structural extrapolation penalty widens as $k$ and model scale increase. This indicates that structural extrapolation might be a bottleneck for standard autoregressive transformers. If the model has not seen the underlying syntactic structures during training, simply sampling more candidates at test time is highly inefficient at discovering them.

*RQ6: Qualitative analysis. What are the characteristics of the generated programs, and what drives generation failure?*

To understand the mechanics of model failure, we analyze the structural and behavioral characteristics of the target programs and the model's corresponding predictions.

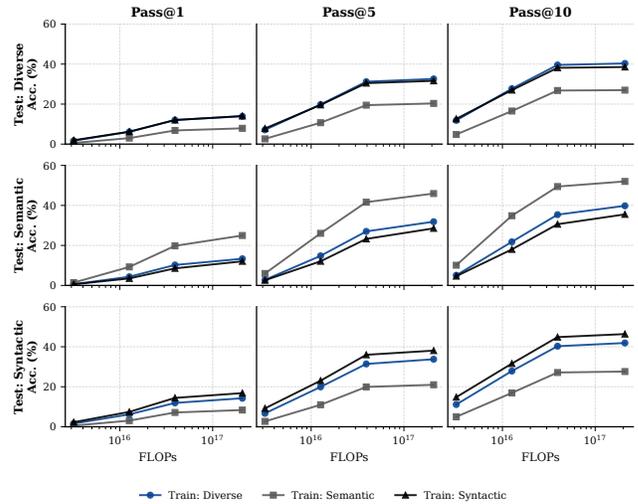

**Figure 9: Density generalization across model and test-time scales.** Performance (pass@1, pass@5, pass@10) of models trained on Diverse, Semantic, and Syntactic distributions, evaluated across corresponding test splits as a function of compute (FLOPs). Increasing the sample budget $k$ (test-time scaling) shifts absolute performance upwards while strictly maintaining the underlying log-linear scaling dynamics. The Diverse training strategy (blue) remains consistently robust across all out-of-distribution shifts regardless of the sampling budget.

**Failure does not correlate with structural complexity.** Figure 11 plots the distribution of solved and failed test instances as a function of the ground-truth program length (measured by the total number of operations in the AST). Across all generalization splits, the distribution of solved cases closely mirrors that of failed cases. Although the models solve a greater number of shorter programs overall, the proportion of solved programs relative to the total number of programs at each length remains largely stable. This pattern is consistent with the hypothesis that the model reflects its training distribution: having been exposed to more short programs, it solves more of them in absolute terms, while maintaining similar success rates across lengths.

**Failures are often good semantic approximations.** Figure 12 investigates the exact nature of the failed generations. First, we note the extreme rarity of syntax errors. Across 1,000 test cases per split, the models produce invalid syntax in only ∼ 20 to 26 instances. This demonstrates that the transformers perfectly internalize the rules of the context-free grammar. Second, when a model fails to perfectly match the target behavior, we observe that its best predictions (the candidate with the lowest mean squared error compared to the target trace) are overwhelmingly clustered at the lower end of the error spectrum (MSE ≈ $10^{-2}$ to $10^{-3}$). Rather than producing random garbage or wildly incorrect functions, the models generate semantic "near-misses." This suggests that the transformer successfully navigates to the correct general behavioral

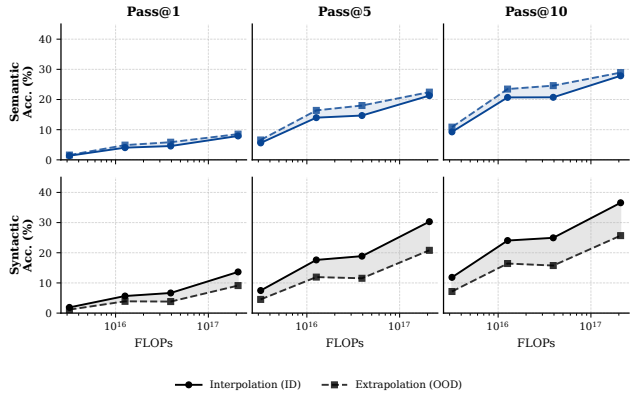

Figure 10: Support generalization across model and test-time scales. Interpolation (solid) versus extrapolation (dashed) performance across semantic (top) and syntactic (bottom) manifolds for pass@1, pass@5, and pass@10. While increasing the test-time sampling budget $k$ improves absolute accuracy, it completely fails to close the persistent out-of-distribution gap (shaded area) on the syntactic manifold. Conversely, semantic extrapolation remains tightly coupled with interpolation across all scales and budgets.

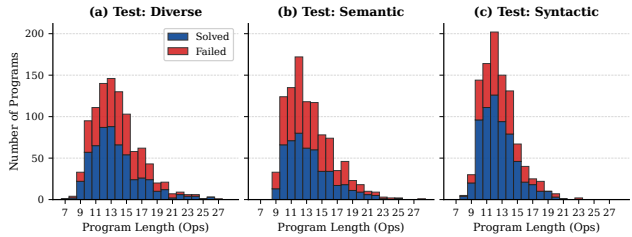

Figure 11: Generalization success by target program length. Distribution of solved (blue) and failed (red) test instances grouped by the true underlying program length (number of operations in the AST). Across all generalization splits, the model's overall success rate is skewed towards shorter programs. Nevertheless, the proportion of failures seems to be roughly consistent across lengths.

neighborhood within the continuous semantic manifold but lacks the precise structural combinatorial reasoning required to close the final gap and achieve exact functional equivalence.

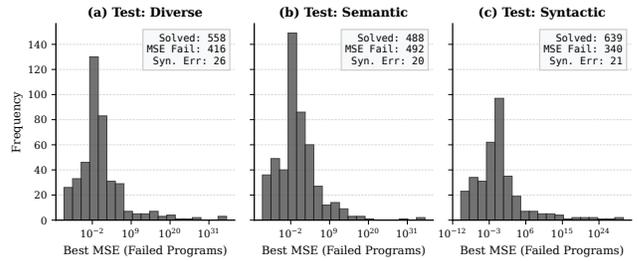

Figure 12: Error analysis of failed generations. Distribution of the minimum mean squared error (MSE) among the generated candidates for unsolved test instances. The text annotations display the number of perfectly solved instances, failures due to incorrect mathematical behavior (MSE Fail), and failures due to generating invalid syntax (Syn. Err). The model successfully internalizes the grammar, almost never generating invalid syntax. Failures are overwhelmingly due to semantic mismatches, many of which cluster at lower errors ($10^{-2}$), representing functional "near misses".

## 4 Semantic Uniqueness Experiment

To measure the maximal generalization ability of the transformer-based setup, we perform a second experiment on a strictly constrained dataset. Instead of enumerating and retaining all syntactically valid alternatives per equivalence class, we retain only a single canonical representative for each unique semantic behavior. The setup is as follows:

*Enumeration and Reduction.* We perform a brute-force enumeration of all valid programs in $\mathcal{G}$ up to length $L$, yielding a raw set of programs $\mathcal{P}_{raw}$. We define an observational equivalence relation $\sim$ over a discrete evaluation domain $\mathcal{D}$. Two programs $p_i, p_j \in \mathcal{P}_{raw}$ are considered semantically equivalent if their outputs are strictly identical for all inputs:

$$p_i \sim p_j \iff \forall x \in \mathcal{D},\ p_i(x) = p_j(x) \tag{8}$$

We partition $\mathcal{P}_{raw}$ into equivalence classes $\mathcal{P}_{raw}/\sim$. For each class, we select the **shortest program** (in token length) as the canonical representative. Our final program universe $\mathcal{U}$ is the set of these unique canonical representatives. This guarantees that every program instance in our downstream dataset embodies a uniquely identifiable functional logic which occurs exactly once in the whole dataset.

### 4.1 Experimental Setup and Design Differences

To rigorously test the limits of our findings, we reconstruct the experiments from the main text under a vastly more difficult condition.

**Differences from the main experiment.** In the primary experiments (Section 4 of the main paper), we retained all syntactically unique alternatives for every equivalence class. Consequently, if a specific mathematical behavior could be computed by five different distinct minimal ASTs, all five were present in the universe $\mathcal{U}$. This allowed the model to learn from much richer structural information.

In this supplementary experiment, we enforce **strict semantic uniqueness**. As described above, we collapse all redundancies and keep only a single canonical representative per functional equivalence class. Therefore, every function in the dataset has exactly one valid syntactic structure in the training data (if it is in-distribution) and strictly zero if it is drawn from an out-of-distribution split.

**Dataset and training.** This aggressive deduplication drastically shrinks the total program universe. Following the same manifold projection, radius partitioning, and sampling strategies as the main paper, we construct new density and support splits (visualized in Figures 13, 14, and 15). Because the universe is smaller, we scale down the training set to 100,000 examples. The test set size is kept the same at 1,000 examples. All model architectures, hyperparameters, and evaluation metrics (*pass@k*) remain identical to those used in the main text.

### 4.2 Results and Analysis

*RQ1: Density generalization. To what extent does model performance degrade when moving out of the training distribution?*

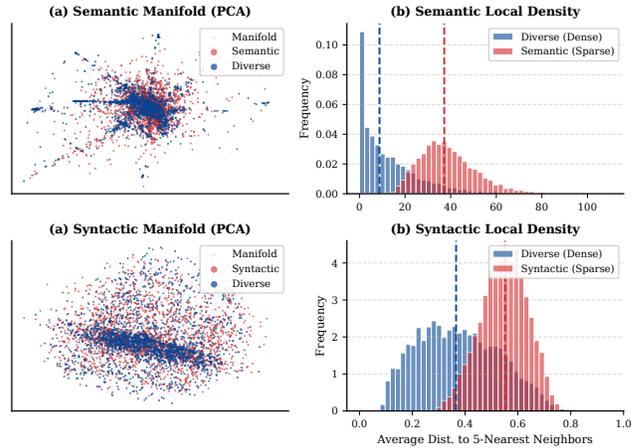

**Figure 13: Visualizing density generalization splits.** Projection of the Diverse, Semantic, and Syntactic datasets onto the continuous manifolds for the strictly deduplicated dataset. The scatter plots (a) and local density histograms (b) confirm that inverse-density sampling (red) successfully targets sparse regions and flattens the density distribution, even after all semantic redundancies have been removed from the search space.

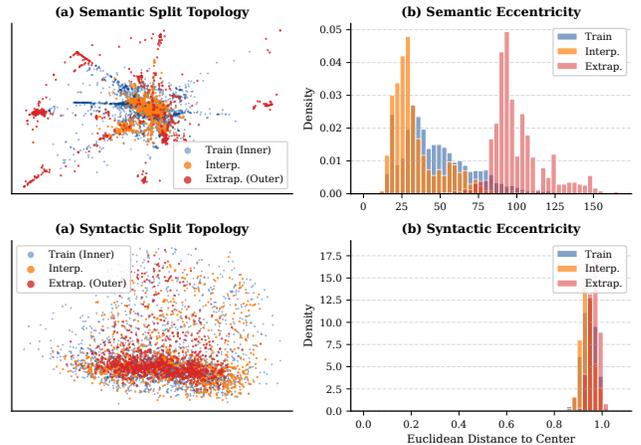

**Figure 14: Visualizing support generalization splits.** Geometric partitioning of the semantically unique program space. The scatter plots (a) and eccentricity histograms (b) demonstrate the strict spatial separation of the interpolation regime (inner radius) from the pure extrapolation regime (outer radius) across both the semantic and syntactic manifolds.

Table 2 reveals a massive drop in absolute performance compared to the main experiment (e.g., peak *pass@1* drops from $\sim$ 30% to $\sim$ 2%). This confirms that the presence of multiple syntactic alternatives in standard datasets acts as a significant crutch for the transformer model. However, the core relative dynamics hold perfectly: the

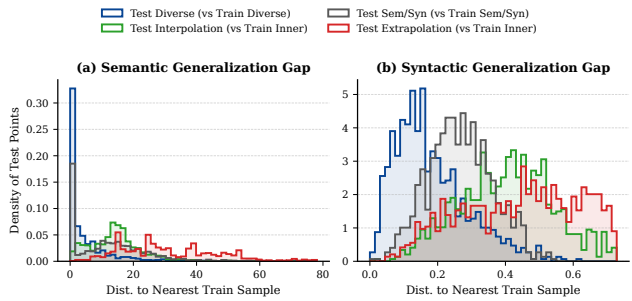

Figure 15: Quantifying out-of-distribution distances. Distribution of test instances based on their exact geometric distance to the nearest training sample. The pronounced rightward shift of the Extrapolation and inverse density sampled (Sem/Syn) test sets confirms their strict out-of-distribution nature, forcing the models to generalize without the crutch of semantically identical syntactic alternatives.

Table 2: Density generalization. Mean and standard deviation across 5 independent runs for $pass@k$ ($k \in \{1, 5, 10\}$). Models trained on specific distributions (rows) are evaluated on test splits (columns). Best out-of-distribution performances for each metric are in bold.

| Train ↓ | Metric | Test Distribution → | | |
|---|---|---|---|---|
| | | Diverse | Semantic | Syntactic |
| Syntactic | pass@1 | 0.006 ± 0.000 | 0.002 ± 0.000 | 0.006 ± 0.000 |
| | pass@5 | 0.024 ± 0.001 | 0.009 ± 0.001 | 0.028 ± 0.001 |
| | pass@10 | 0.041 ± 0.002 | 0.017 ± 0.002 | 0.048 ± 0.002 |
| Semantic | pass@1 | 0.004 ± 0.000 | **0.024 ± 0.001** | 0.004 ± 0.000 |
| | pass@5 | 0.016 ± 0.001 | **0.087 ± 0.003** | 0.013 ± 0.001 |
| | pass@10 | 0.028 ± 0.002 | **0.133 ± 0.004** | 0.022 ± 0.002 |
| Diverse | pass@1 | **0.008 ± 0.000** | 0.005 ± 0.000 | **0.007 ± 0.000** |
| | pass@5 | **0.035 ± 0.001** | 0.021 ± 0.001 | **0.032 ± 0.001** |
| | pass@10 | **0.061 ± 0.003** | 0.037 ± 0.002 | **0.057 ± 0.002** |

**Diverse** training strategy remains the most robust against out-of-distribution shifts, maintaining stable (albeit low) performance across all test splits. In contrast, models trained strictly on a single manifold (Semantic or Syntactic) suffer severe degradation when evaluated out-of-distribution.

*RQ2: Support generalization. Can the model extrapolate to novel structures and functional behaviors outside the support of its training data?*

Table 3 strongly reinforces our findings regarding extrapolation. When evaluating on unseen functional behaviors (semantic extrapolation), the model does not suffer a relative drop; it even slightly improves relative to its interpolation baseline. Conversely, extrapolating to unseen syntactic structures triggers an immense performance collapse, with relative metrics dropping by over 55%. This continues to indicate that when forced to learn a mapping of syntax to semantics, transformers can extrapolate behaviorally using known structures, but fail to invent novel structural combinations.

*RQ3: Scaling behavior. How is generalization performance influenced by model scale?*

As shown in Figure 16, scaling compute yields continuous improvements, but the trajectory remains strictly log-linear. Furthermore, Figure 16e demonstrates that the massive gap between syntactic interpolation and extrapolation persists across all model scales. Removing semantic redundancies does not alter the architectural bottleneck: exponential compute is required for linear gains on strictly OOD problem instances.

Table 3: Support generalization. Performance gap when transitioning from interpolation (within training support) to pure extrapolation (outside support) evaluated across $pass@k$ over 5 runs.

| Manifold | Metric | Generalization Regime | | Gap |
|---|---|---|---|---|
| | | Interpolation | Extrapolation | |
| Semantic | pass@1 | 0.001 ± 0.000 | 0.002 ± 0.000 | +19.078% |
| | pass@5 | 0.007 ± 0.001 | 0.008 ± 0.001 | +20.392% |
| | pass@10 | 0.013 ± 0.001 | 0.016 ± 0.001 | +21.426% |
| Syntactic | pass@1 | 0.004 ± 0.000 | 0.002 ± 0.000 | −55.712% |
| | pass@5 | 0.020 ± 0.002 | 0.009 ± 0.001 | −55.834% |
| | pass@10 | 0.037 ± 0.002 | 0.016 ± 0.001 | −55.954% |

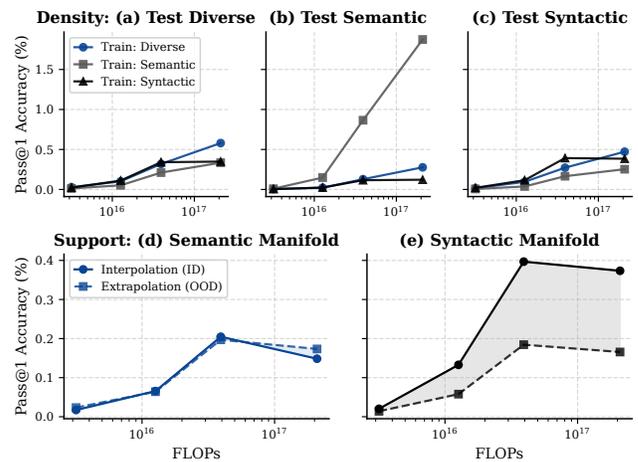

Figure 16: Scaling laws for generalization. Performance ($pass@1$) as a function of compute (FLOPs). (a–c) Density generalization: All models exhibit strictly log-linear improvements regardless of the sampling strategy. (d–e) Support generalization: Semantic extrapolation (d) scales identically to interpolation, whereas structural extrapolation (e) suffers a persistent OOD penalty that cannot be easily overcome by simply scaling up compute.

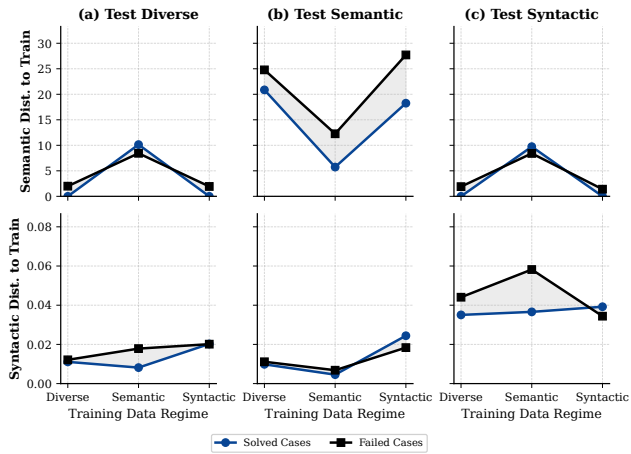

Figure 17: Distance to nearest training neighbor. Comparing distance of solved (blue) vs. failed (black) test instances across semantic (top) and syntactic (bottom) manifolds. When evaluated out-of-distribution (panels b and c), models successfully solve instances that are significantly closer to their pre-training support, indicating a reliance on local manifold interpolation.

*RQ4: Qualitative analysis.* What geometric conditions dictate whether a model successfully generalizes to a novel instance?

Figure 17 maps the distance to the nearest training neighbor for solved vs. failed instances. Mirroring the main paper, successful OOD evaluations (solved cases in panels b and c) are on average closer to the training support than failed instances. This hints that also in a strictly deduplicated program space, transformer generalization relies on local geometric interpolation from known training anchors.

*RQ5: Test-time scaling.* How is generalization performance influenced by test-time scaling, i.e., the number of sampled candidates $k$?

**Density generalization.** Figure 18 tracks density generalization across different sampling budgets ($k \in \{1, 5, 10\}$). While generating more candidates naturally raises absolute performance across the board, the underlying log-linear scaling curves and relative model rankings remain completely rigid. The Diverse model maintains its robust OOD superiority regardless of how many beams are sampled.

**Support generalization.** Figure 19 evaluates test-time scaling for extrapolation. Crucially, the syntactic extrapolation gap (the shaded area in the bottom row) cannot be closed by simply increasing $k$. Generating more candidate programs helps the model find solutions within the interpolation regime, but does almost nothing to help it guess structurally novel AST combinations outside its pre-training support.

*RQ6: Qualitative analysis.* What are the characteristics of the generated programs, and what drives generation failure?

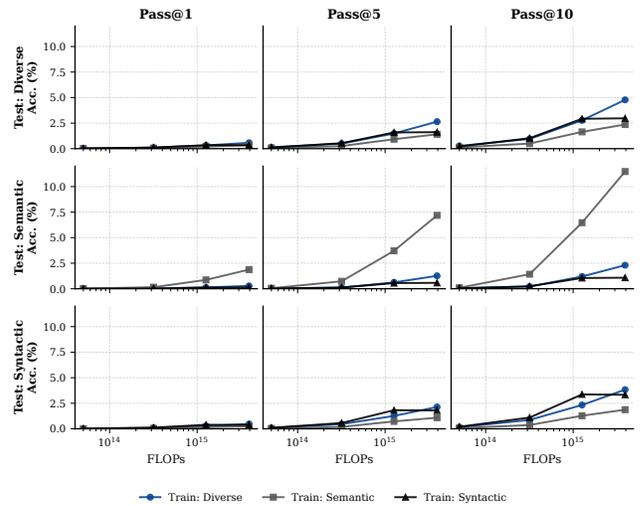

Figure 18: Density generalization across scales. Performance (pass@k) of models evaluated across data distributions as a function of compute. Because removing semantic ambiguity makes the synthesis task vastly more difficult, absolute accuracy is drastically reduced compared to the main experiment. However, the fundamental log-linear scaling dynamics and the relative robustness of the Diverse training strategy remain entirely consistent.

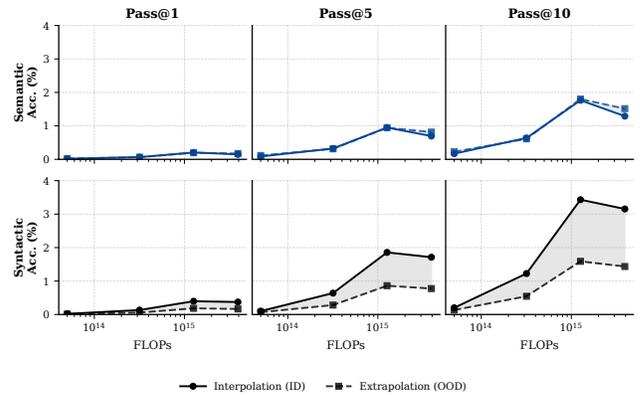

Figure 19: Support generalization across scales. Interpolation versus extrapolation performance across semantic (top) and syntactic (bottom) manifolds. The structural extrapolation penalty (shaded area, bottom row) remains severe and widens with increased scale and sampling budgets, confirming that test-time scaling cannot overcome the model's limitations to synthesize unseen syntactic structures.

**Length and complexity.** Figure 20 illustrates the extreme difficulty of the semantic uniqueness constraint. A much smaller amount of problems overall is solved, in tendency shorter programs are solved to a higher amount than longer ones.

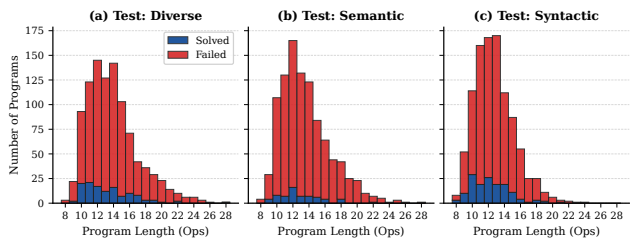

Figure 20: Generalization success by program length. Distribution of solved and failed instances grouped by the ground-truth AST length. The removal of semantic redundancies vastly increases the overall failure rate (red). In tendency, failures increase as target programs become structurally longer and compositionally more complex.

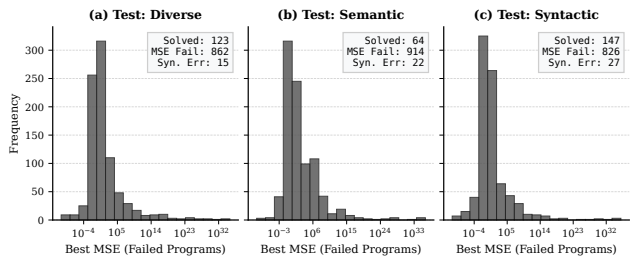

Figure 21: Error analysis of failed generations. Distribution of the minimum mean squared error (MSE) among generated candidates. Even on this highly constrained dataset, the models rarely produce invalid syntax (Syn. Err.). Failures remain predominantly semantic "near-misses," clustering at lower error magnitudes ($10^{-3}$ to $10^{-6}$) where the model approximates the target behavior without achieving perfect functional equivalence.

**Error distributions.** Figure 21 confirms that the models still learn the grammar perfectly; syntactic validity errors (Syn. Err.) remain negligible. Furthermore, the MSE of failed programs still clusters at the lower end of the error spectrum ($10^{-3}$ to $10^{-6}$). The Model frequently generates structurally valid program that are semantically very close to the target function. It often fails to nail the exact canonical expression required to pass the strict functional equivalence test.

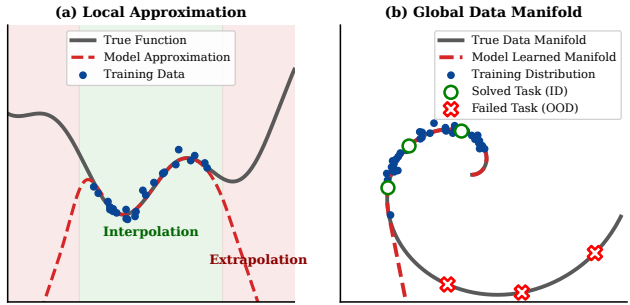

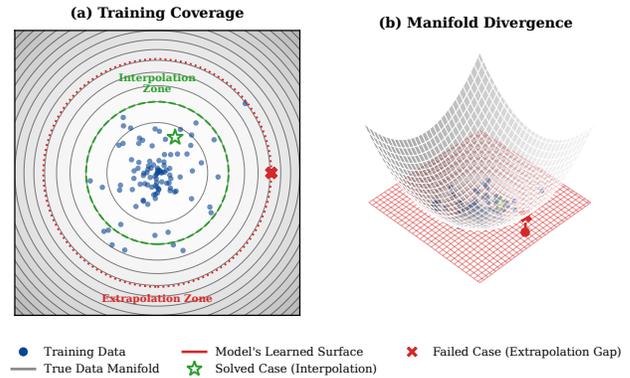

Figure 22: Manifold approximation (1D Example). (a) Local approximation: The model successfully fits a curve (dashed red) to the training data (blue dots) within the interpolation regime (green). Near the boundary, the approximation may hold briefly, allowing for "near-OOD" extrapolation. However, as it moves further into the pure extrapolation regime (red), the approximation breaks down entirely, diverging from the true function (solid gray). (b) Global data manifold: Looking at it globally, the model only learns the local curvature of the manifold where data is present. Solved OOD tasks (green circles) occur slightly outside the training bounds where the learned trajectory still aligns with the true manifold, whereas failed tasks (red crosses) occur where the true manifold diverges into unseen geometry.

## 5 Discussion: Manifold Approximation

To explain the empirical behaviors observed in our controlled experiments—specifically the strict log-linear scaling bottlenecks, the tight correlation between nearest-neighbor proximity and success, and the severe penalty of structural extrapolation—we discuss potential explanations for this behavior. We posit that autoregressive transformers do not inherently learn decomposable building blocks and assemble them from first principles to solve novel problems. Instead, they seem to learn a continuous, high-dimensional approximation of the data manifold defined by their pre-training distribution. Generalization, in this representation, becomes a measure of geometric proximity: the model successfully synthesizes solutions for novel problem instances if and only if they fall within, or very close to, the local geometry it has explicitly learned to approximate. This aligns with findings by [2], who demonstrated that transformers rely on local pattern matching and struggle with true compositional extrapolation on multi-step reasoning tasks.

**Local approximation.** As conceptually illustrated in Figure 22a, we hypothesize that the model fits a local surface to the discrete training data, akin to a high-dimensional spline or polynomial regression. Within the interpolation regime, defined by the support of the training data, the model is highly constrained by surrounding data anchors, allowing for robust and accurate synthesis. As a query moves just beyond the boundary of the training data, the local gradient or curvature learned at the edge of the distribution may remain valid for a short distance. This explains why the model occasionally solves extrapolation tasks that possess a very close nearest neighbor in the training set. However, as the target moves

Figure 23: Manifold approximation (2D Example). (a) Training coverage: A top-down view showing dense training data in the center. The interpolation zone covers the area where data density is high, making successful synthesis highly probable (green star). (b) Manifold divergence: A 3D view showing the model's learned flat plane (red grid) approximating the true non-linear data manifold (gray surface) locally. In the center, the approximation is highly accurate. However, as one moves into the extrapolation zone, the true manifold curves upward while the model's learned approximation continues linearly, causing a massive divergence that results in generalization failure (red cross).

further into the pure extrapolation regime, the unconstrained approximation diverges wildly from the true underlying function, leading to generalization failure.

**Global data manifold.** When projecting this concept onto a global structure (Figure 22b), the limitations of pure neural generation become apparent. Assuming the target programs exist on a highly complex, non-linear true data manifold. The model only learns the specific sub-manifold where it receives a dense learning signal. When forced to extrapolate to entirely novel behaviors or structures, the model simply extends its learned local geometry into unexplored space. If the true data manifold sharply curves or exhibits novel geometric features which cannot be explained or mapped to known syntactic structures, the model's approximation completely misses the target manifold, resulting in the relatively high absolute syntactic and semantic error rates, respectively, the low to modest and log-linearly increasing generalization performance observed during out-of-distribution evaluation.

**Training coverage.** Figure 23 illustrates this in another example. In the center, the true non-linear data manifold (the gray bowl) is reasonably approximated by the model's learned flat plane (the red grid). For test points within this dense interpolation zone (green star), the approximation gap is minimal, and the model succeeds. However, in the extrapolation zone, the true manifold curves upward while the model's approximation continues flatly, creating an insurmountable divergence gap (red cross). The example illustrates the importance of a widespread coverage of the data manifold via appropriate training data selection. By diversely sampling data across

the entire manifold during pre-training, more accurate structural "anchor points" across the wider curvature of the true manifold can be learned by the model. This forces the model's approximation surface to bend and conform to the true global geometry, minimizing the divergence gap globally and thereby maximizing out-of-distribution robustness.



\end{document}